\documentclass[pdflatex,sn-mathphys-num]{sn-jnl}

\usepackage{graphicx}%
\usepackage{multirow}%
\usepackage{amsmath,amssymb,amsfonts}%
\usepackage{amsthm}%
\usepackage{mathrsfs}%
\usepackage[title]{appendix}%
\usepackage{xcolor}%
\usepackage{textcomp}%
\usepackage{manyfoot}%
\usepackage{booktabs}%
\usepackage{algorithm}%
\usepackage{algorithmicx}%
\usepackage{algpseudocode}%
\usepackage{listings}%
\usepackage{hyperref}
\usepackage{multirow}
\usepackage{url}
\usepackage{array}
\usepackage{multirow}
\newcolumntype{L}[1]{>{\raggedright\arraybackslash}p{#1}}
\usepackage{xcolor}
\usepackage{tcolorbox}
\usepackage{tabularx}
\usepackage{adjustbox}
\usepackage{booktabs}
\usepackage{pdflscape}
\usepackage{graphicx}
\usepackage{rotating}
\usepackage{array}
\usepackage{multirow}
\usepackage{xurl}
\usepackage{pdflscape}
\usepackage{geometry}
\usepackage{svg}
\usepackage{graphicx}
\usepackage{pgfplots}
\usepackage{tikz}
\usepackage{xcolor}
\usetikzlibrary{arrows.meta,positioning,fit,backgrounds}
\pgfplotsset{compat=1.18}

\theoremstyle{thmstyleone}%
\theoremstyle{thmstyletwo}%

\theoremstyle{thmstylethree}%

\begin{document}

\title{Agentic Frameworks for Reasoning Tasks: An Empirical Study}


\author*[1,2]{\fnm{Zeeshan} \sur{Rasheed}}\email{zeeshan.rasheed@tuni.fi}

\author[2,3]{\fnm{Abdul} \sur{Malik Sami,}}\email{malik.sami@tuni.fi}

\author[1,2]{\fnm{Muhammad} \sur{Waseem}}\email{muhammad.waseem@tuni.fi}

\author[1,2]{\fnm{Kai-Kristian} \sur{Kemell}}\email{kai-kristian.kemell@tuni.fi}

\author[1,2]{\fnm{Mika} \sur{Saari}}\email{mika.saari@tuni.fi}

\author[]{\fnm{Pekka} \sur{Abrahamsson}}\email{pekka.abrahamsson@tuni.fi}



\affil*[]{\orgdiv{Faculty of Information Technology and Communication Sciences}, \orgname{Tampere University}, \orgaddress{\city{Tampere}, \country{Finland}}}




\abstract{Recent advances in agentic frameworks have enabled the development of AI agents capable of complex reasoning and decision-making. However, systematic evidence comparing their reasoning performance, efficiency, and real-world suitability remains limited, making it difficult to select appropriate frameworks for practical use. To address this gap, we empirically evaluate 22 widely used agentic frameworks across three reasoning benchmarks: BBH, GSM8K, and ARC. We selected these frameworks from 1,200 GitHub repositories collected between January 2023 and July 2025, and developed a taxonomy based on their architectural designs. We then evaluated these frameworks under a unified experimental setting, measuring reasoning accuracy, execution time, computational cost, and the consistency of framework performance across the benchmarks.

Our results show that 19 of the 22 frameworks completed all three benchmarks. Among these, 12 frameworks showed consistent performance, with a mean accuracy of 74.6–75.9\%, execution time of 4–6 seconds per task, and cost of 0.14–0.18¢ per task. The remaining frameworks performed worse, primarily because of orchestration issues rather than limitations in reasoning. For instance, Camel failed to complete BBH after 11 days of runtime because of uncontrolled context growth; Upsonic consumed \$1,434 in a single day because extraction failures led to multiple retries, while uncontrolled context growth sharply increased prompt-token usage; and AutoGen and Mastra exhausted API quotas through iterative agent interactions that increased prompt length without improving the answers. These are system-level failures in memory management, retry policy, and context handling rather than failures in reasoning.
We further found that all frameworks, including the high performers, degrade sharply regarding mathematical reasoning: among the frameworks that completed all benchmarks, mean accuracy on GSM8K was 44.35\%, compared with 89.80\% on BBH and 89.56\% on ARC. This gap is consistent across architectures, reinforcing that current agentic designs inherit rather than overcome the base model’s weaknesses in multi-step numerical computation. 

Overall, this study provides a systematic comparison and architectural perspective on agentic frameworks to support framework selection for reasoning-intensive software engineering tasks. Our results suggest that such selection should be guided primarily by orchestration quality—particularly memory discipline, failure handling, and cost control—rather than by architectural category or claimed reasoning capabilities alone. We have released the dataset, source code, and per-task logs to support replication and further research.}


\maketitle

\section{Introduction}\label{sec1}
In recent years, AI agents have received increasing attention in relation to performing complex tasks that require planning, reasoning, and tool usage \cite{liu2024large, he2025llm}.  
At the same time, recent developments in agentic frameworks have improved agents’ ability to design multi-agent workflows, coordinate specialized roles, and execute complex tasks autonomously \cite{vaidhyanathan2025agentic}.
According to Wang \textit{et al.} \cite{wang2025empirical}, agentic frameworks are toolkits or libraries that provide reusable components, along with built-in coordination and memory capabilities, to support the design, deployment, and orchestration of autonomous or semi-autonomous agents. The growing adoption of these frameworks in both academia and industry demonstrates their increasing importance in agent-based system development \cite{zhao2025llm}.

Reasoning is a core capability of intelligent agents, enabling them to perform logical inference, solve problems, and make decisions in dynamic and interactive environments \cite{wei2026agentic}. As a result, agentic frameworks have been explored for applications that require advanced reasoning capabilities \cite{zhao2025llm, lu2025octotools}. 
However, despite their widespread use, there is still a lack of comprehensive empirical studies that systematically evaluate and compare agentic frameworks in terms of reasoning performance, efficiency, and practical effectiveness in software engineering contexts \cite{wei2026agentic, wang2025empirical}. Addressing this gap is essential for understanding the strengths and limitations of existing frameworks and supporting more informed framework selection in practice.



\textbf{Motivation}: This study is part of the MAISA project (2024–2026) and the ANSE project (2025–2027), both funded by Business Finland. These projects bring together academic researchers and major Finnish companies to explore the use of agentic AI and large language model (LLM)-based technologies to advance software engineering practices. This collaboration focuses on developing, assessing, and facilitating the adoption of innovative AI-driven approaches in real-world software engineering environments.
This study aims to address industrial needs by systematically evaluating the capabilities of agentic frameworks for reasoning tasks and examining their practical trade-offs in software engineering contexts. Recent studies \cite{wang2025empirical, garg2025designing} highlight that, despite the rapid emergence of these frameworks, industry practitioners still lack clear guidance on which ones are most suitable for different application contexts, making it difficult to assess trade-offs related to reliability, scalability, and reasoning performance. 
As several studies \cite{kazemi2025big, suzgun2022challenging, yu2024humaneval, frendi2026comprehensive, hasan2026assessing} have conducted benchmark-based empirical studies to evaluate and compare LLM capabilities across various software engineering tasks, the growing practical adoption of LLM-based agents signals the need for more comprehensive benchmarking of agentic frameworks \cite{wei2026agentic}. Although several studies \cite{yin2025comprehensive, lu2025octotools, barbarroxa2024benchmarking, serafim2025comparative} have compared such frameworks for different software engineering tasks using specific benchmarks and evaluation metrics, these studies typically consider only a limited number of frameworks or focus on narrow aspects of performance. As a result, there remains a lack of broad empirical evidence that can support framework selection for reasoning-intensive software engineering applications.

To this end, this study addresses the needs of industrial practitioners by providing a comprehensive empirical evaluation of agentic frameworks, enabling a structured comparison across key dimensions and supporting informed decision-making for their practical adoption. In this work, we evaluate 22 agentic frameworks (see Table \ref{tab:agentic_taxonomy}), selected through a systematic process, and develop a taxonomy based on their architectural design. We categorize these frameworks into five architectural types: single-agent, role-based multi-agent, hierarchical, modular, and graph-based architectures. We also identify their adopted reasoning strategies and memory mechanisms.
These 22 frameworks are empirically assessed in a unified experimental setting to compare their reasoning performance, trade-offs, and consistency across three widely recognized reasoning benchmarks: Big-Bench Hard (BBH) \cite{suzgun2022challenging}, GSM8K \cite{cobbe2021training}, and the AI2 Reasoning Challenge (ARC) \cite{clark2018think}. These benchmarks were selected because they contain diverse datasets designed to evaluate the complex reasoning capabilities of AI systems.

\textbf{Contributions}: The key contributions of this paper are summarized as follows:
\begin{itemize}

\item Systematic selection of agentic frameworks and creation of a taxonomy based on their architectural design, providing an overview of the current framework landscape.

\item Development of a unified experimental setting for comparing the reasoning accuracy, response time, operational cost, and consistency of agentic frameworks across benchmarks, thereby providing insights into the computational trade-offs associated with their deployment as well as their performance consistency.

\item Empirical insights into the capabilities of agentic frameworks for reasoning-intensive tasks, offering practical guidance for researchers and industry practitioners in real-world applications.

\item Public release of the dataset and source code to support the replication, validation, and extension of the study. The source code is available via GitHub \cite{agenticframeworkcomparison} and the dataset, including the analyzed frameworks, benchmark results, and supporting analysis, is available on Zenodo \cite{rasheed_2026_19593242}, thereby promoting transparency, reproducibility, and future research on agentic frameworks.

\end{itemize}

\textbf{Structure of the paper:} 
The work related to this study is presented in Section~\ref{Related Work}, followed by the study design in Section~\ref{Study Design}. The results of the empirical analysis are reported in Section~\ref{Result}. The key findings and their implications are discussed in Section~\ref{Discussion}, while threats to validity are examined in Section~\ref{Threats to Validity}. Finally, Section~\ref{conclusion} summarizes the paper and outlines directions for future research.

\section{Related Work}
\label{Related Work}

This section first provides an overview of agentic frameworks in Section~\ref{Agentic Frameworks: Architectures and Capabilities}, outlining their coordination strategies and workflow management approaches. In Section~\ref{Agentic Framework Evaluation} we then examine prior studies that evaluate agentic frameworks across various tasks.

\subsection{Agentic Frameworks: Architectures and Capabilities} 
\label{Agentic Frameworks: Architectures and Capabilities}



LLMs have delivered strong performance across a broad range of tasks under direct prompting, establishing single-model interaction as an important baseline \cite{aratchige2025llms, rasheed2026llm}. However, complex reasoning and software engineering tasks often require iterative planning, task decomposition, tool use, memory, and coordination across multiple steps \cite{sun2025llm}. Intelligent agents address these challenges by enabling LLMs to operate through structured, goal-directed workflows rather than isolated prompt-response interactions \cite{he2025llm}. Prior to the development of agentic frameworks, traditional agent development required the implementation of orchestration logic, state management, communication protocols, and execution pipelines from scratch, which demanded high levels of technical expertise \cite{abou2025agentic}. Moreover, these approaches offered limited support for external tool integration and persistent memory, thereby constraining the development of scalable and context-aware intelligent agents \cite{bandi2025rise,zhang2025survey}. Agentic frameworks were introduced to address these limitations by providing reusable abstractions and built-in support for agent orchestration, memory management, tool integration, and multi-step execution, thereby simplifying the development and deployment of intelligent agent systems \cite{wang2024agent}.

Since 2023, the development of agentic frameworks such as AutoGen \cite{wu2024autogen}, Camel \cite{li2023camel}, CrewAI \cite{crewai2025}, SuperAGI \cite{superagi2025}, TaskWeaver \cite{qiao2023taskweaver}, MetaGPT \cite{hong2023metagpt}, and ChatDev \cite{qian2024chatdev} has simplified the development of intelligent agent systems. These frameworks support complex functionalities, including state management, tool integration, and inter-agent communication \cite{vaidhyanathan2025agentic}. By providing structured APIs for defining agent roles, behaviors, and collaborative workflows, they enable more efficient development of advanced single-agent and multi-agent applications \cite{handler2023balancing}.

These frameworks organize agentic systems in different ways to support collaboration and task execution. In particular, they adopt a variety of architectural designs for agent workflows, including single-agent architectures, role-based multi-agent architectures, hierarchical architectures, modular architectures, and graph-based architectures \cite{patel2025systematic}, \cite{qiao2023taskweaver}, \cite{vaidhyanathan2025agentic}.
Single-agent architectures rely on a centralized agent responsible for planning, reasoning, and executing tasks \cite{qiao2023taskweaver}. While this approach simplifies coordination, it may face limitations when handling complex and large-scale tasks. Frameworks such as TaskWeaver \cite{qiao2023taskweaver}, AutoGPT \cite{yang2023auto}, LangChain \cite{langchain2023}, and LlamaIndex \cite{llamaindex2023} adopt this architectural design \cite{qiao2023taskweaver}. In addition, frameworks such as LangChain and LlamaIndex further extend agentic systems by supporting interaction with external environments, large document collections, and complex execution flows. These frameworks provide abstractions for chaining reasoning steps, integrating external knowledge through retrieval-augmented generation, and managing agent memory \cite{topsakal2023creating}, \cite{wang2025empirical}.

In contrast, role-based multi-agent architectures structure collaboration by assigning specialized roles and responsibilities to different agents, enabling coordinated problem-solving through defined interaction protocols \cite{hasan2025empirical}. This approach is followed by frameworks such as AutoGen \cite{wu2024autogen}, Camel \cite{li2023camel}, CrewAI \cite{crewai2025}, and SuperAGI \cite{superagi2025}. These frameworks often incorporate explicit memory management mechanisms—such as short-term conversational memory, shared context buffers, and persistent storage—to maintain state across multi-step interactions \cite{wei2026agentic}.

Hierarchical architectures introduce a layered organization of agents, typically consisting of high-level planning agents and lower-level execution agents \cite{patel2025systematic}. This structure facilitates effective task decomposition, coordination, and management of complex workflows. Frameworks such as MetaGPT \cite{hong2023metagpt} and Google ADK \cite{google_adk_python2026} exemplify this architectural paradigm \cite{vaidhyanathan2025agentic}.
Modular architectures focus on decomposing agent systems into reusable and interchangeable components, such as planning, memory, and tool-use modules \cite{patel2025systematic}. This design enhances the extensibility, maintainability, and customization of agentic workflows. Frameworks such as Mastra represent this architectural approach \cite{vaidhyanathan2025agentic}.

Graph-based architectures model agent workflows as directed graphs, where nodes represent agents or tasks and edges define dependencies and execution flow \cite{patel2025systematic}. This enables dynamic, flexible, and stateful orchestration of complex processes. LangGraph is a prominent example of this architectural paradigm, which enables more reliable and flexible coordination of agent behavior across complex workflows \cite{patel2025systematic}.

\subsection{Agentic Framework Evaluation}
\label{Agentic Framework Evaluation}
In recent years, several studies have been conducted to evaluate the suitability of agentic frameworks for various software engineering tasks. As summarized in Table~\ref{tab:agentic_framework_eval}, existing studies compare agentic frameworks using five different approaches: (1) benchmark-based evaluation, (2) metrics-based evaluation, (3) architecture-based evaluation, (4) qualitative analysis of GitHub discussions, and (5) survey-based evaluation. 

\textbf{Benchmark-based evaluation}:
Several recent studies evaluate the performance of agentic frameworks using benchmark-driven experiments. Yin \textit{et al}. \cite{yin2025comprehensive} compare seven agentic frameworks across three widely adopted benchmarks designed for software engineering tasks. These include the SRDD benchmark, the LLM-SmartAudit benchmark for vulnerability detection, and the SWE-bench Lite benchmark for program repair. Their study analyzes the capability of different frameworks in identifying code vulnerabilities and performing automated software maintenance tasks. 
Similarly, Liu \textit{et al}. \cite{lu2025octotools} propose the OctoTools framework for solving complex reasoning tasks and evaluate it against three existing frameworks—AutoGen, LangChain, and GPT-Function. The comparison is conducted across sixteen reasoning benchmarks, where the proposed OctoTools framework demonstrates comparatively higher accuracy on several tasks. 

In addition, Barbarroxa \textit{et al}. \cite{barbarroxa2024benchmarking} compare three multi-agent frameworks: AutoGen, CrewAI, and TaskWeaver, using a machine learning code-generation case study. In their evaluation, each framework generates code to build energy forecasting models from a shared dataset, and the resulting models are assessed on a held-out dataset using the RMSE metric. This study provides a direct task-level comparison of agentic frameworks and complements research that focuses primarily on development practices rather than empirical performance.

\textbf{Metrics-based evaluation}:
Two studies compare agentic frameworks using task-specific evaluation metrics to assess their effectiveness. For instance, Serafim and Mariama \cite{serafim2025comparative} compare four frameworks: AutoGen, AutoGPT, Dify, and Semantic-kernel, on an optimization task based on the Metagente pipeline, evaluating summarization quality using ROUGE scores. Their results show that no single framework performs best across all dimensions: Dify achieves the highest ROUGE scores, whereas AutoGen and Semantic-kernel demonstrate stronger orchestration flexibility. In contrast, AutoGPT exhibits higher execution time and less efficient agent communication. 
Similarly, Barbarroxa \textit{et al}. \cite{barbarroxa2024benchmarking} utilize the RMSE metric to compare three multi-agent frameworks: AutoGen, CrewAI, and TaskWeaver, in a machine learning code-generation case study, where the generated models are evaluated based on their predictive accuracy.

\begin{table*}[t]
\centering
\caption{Comparison of existing studies on agentic framework evaluation.}
\label{tab:agentic_framework_eval}
\small
\renewcommand{\arraystretch}{1.2}

\resizebox{\textwidth}{!}{
\begin{tabular}{|c|l|l|c|c|c|c|c|}
\hline
\textbf{S.No} & \textbf{Paper} & \textbf{Frameworks} & \textbf{Benchmark} & \textbf{Metrics} & \textbf{Architecture} & \textbf{Survey} & \textbf{Qualitative Analysis} \\
\hline

1 & Yin \textit{et al}. \cite{yin2025comprehensive} & 7 frameworks & SRDD, LLM-SmartAudit, SWE-bench Lite & $\times$ & $\times$ & $\times$ & $\times$ \\
\hline

2 & Liu \textit{et al}. \cite{lu2025octotools} & OctoTools, AutoGen, LangChain, GPT-Function & 16 reasoning benchmarks & Accuracy & $\times$ & $\times$ & $\times$ \\
\hline

3 & Barbarroxa \textit{et al}. \cite{barbarroxa2024benchmarking} & AutoGen, CrewAI, TaskWeaver & $\times$ & RMSE & $\times$ & $\times$ & $\times$ \\
\hline

4 & Serafim and Mariama \cite{serafim2025comparative} & AutoGen, AutoGPT, Dify, Semantic-kernel & $\times$ & ROUGE & $\times$ & $\times$ & $\times$ \\
\hline

5 & Wei \textit{et al}. \cite{wei2026agentic} & 6 frameworks & $\times$ & $\times$ & $\checkmark$ & $\checkmark$ & $\times$ \\
\hline

6 & Aratchige and Ilmini \cite{aratchige2025llms} & AutoGen, Camel, CrewAI, MetaGPT, LangGraph & $\times$ & $\times$ & $\checkmark$ & $\checkmark$ & $\times$ \\
\hline

7 & Guo \textit{et al}. \cite{guo2024large} & MetaGPT, Camel, AutoGen & $\times$ & $\times$ & $\checkmark$ & $\checkmark$ & $\times$ \\
\hline

8 & Ferrag \textit{et al}. \cite{ferrag2025llm} & LangChain, LlamaIndex, CrewAI, Swarm & $\times$ & $\times$ & $\times$ & $\checkmark$ & $\times$ \\
\hline

9 & Bandi \textit{et al}. \cite{bandi2025rise} & Multiple frameworks & $\times$ & $\times$ & $\checkmark$ & $\checkmark$ & $\times$ \\
\hline

10 & Vaidhyanathan \textit{et al}. \cite{vaidhyanathan2025agentic} & 10 frameworks & $\times$ & $\times$ & $\checkmark$ & $\times$ & $\times$ \\
\hline

11 & Patel \textit{et al}. \cite{patel2025systematic} & 6 frameworks & $\times$ & $\times$ & $\checkmark$ & $\times$ & $\times$ \\
\hline

12 & Shi \textit{et al}. \cite{shi2025comparative} & 8 frameworks & $\times$ & $\times$ & $\checkmark$ & $\times$ & $\times$ \\
\hline

13 & Hasan \textit{et al}. \cite{hasan2025empirical} & 39 frameworks & $\times$ & $\times$ & $\times$ & $\times$ & $\checkmark$ \\
\hline

14 & Sun \textit{et al}. \cite{sun2025llm} & 10 frameworks & $\times$ & $\times$ & $\times$ & $\times$ & $\checkmark$ \\
\hline

15 & Wang \textit{et al}. \cite{wang2025empirical} & 10 frameworks & $\times$ & $\times$ & $\times$ & $\times$ & $\checkmark$ \\
\hline

\end{tabular}
}

\end{table*}

\textbf{Survey-based evaluation}:
Several studies compare agentic frameworks through survey-based analyses that examine their architectural characteristics, design principles, and application domains. Wei \textit{et al}. \cite{wei2026agentic} provide a comparative overview of multiple frameworks, including AgentOrchestra, OWL, SE-Agent, Trae, GPTSwarm, OpenHands, and SWE-Agent. Their study analyzes architectural designs, coordination mechanisms, and application domains, highlighting how these frameworks support agent orchestration, task decomposition, tool integration, and inter-agent communication.
Similarly, Aratchige and Ilmini \cite{aratchige2025llms} review frameworks such as AutoGen, Camel, CrewAI, MetaGPT, and LangGraph, outlining the technological landscape of agentic systems while identifying challenges such as coordination overhead and scalability. 

Guo \textit{et al}. \cite{guo2024large} present an overview of multi-agent frameworks and their domain applications, identifying key components of AI agents such as environment interfaces, profiling, communication, and capability acquisition. Their study discusses representative frameworks including MetaGPT, Camel, and AutoGen. Likewise, Ferrag \textit{et al}. \cite{ferrag2025llm} analyze LLM-based agent frameworks across different domains, comparing systems such as LangChain, LlamaIndex, CrewAI, and Swarm in terms of workflows, components, and core design ideas.
Furthermore, Bandi \textit{et al}. \cite{bandi2025rise} conduct a comprehensive survey of agentic systems, summarizing framework architectures, evaluation methods, application domains, and open research challenges.

\textbf{Architecture-based evaluation}:
Several studies compare agentic frameworks by analyzing their architectural designs and system components. Vaidhyanathan \textit{et al}. \cite{vaidhyanathan2025agentic} provide an overview of ten agentic frameworks, examining their architectural patterns, planning and reasoning mechanisms, and memory management strategies used to support agent workflows. 
Similarly, Patel \textit{et al}. \cite{patel2025systematic} compare six popular frameworks: AutoGen, Google ADK, CrewAI, LlamaIndex, LangGraph, and Semantic-kernel, with a focus on their architectural characteristics. Their study proposes a taxonomy of framework design patterns, demonstrates agentic workflows for academic tasks, and discusses open challenges for future research.

Shi \textit{et al}. \cite{shi2025comparative} conducted a comparative analysis of eight agentic frameworks, including LangGraph, CrewAI, OpenAI Swarm, AutoGen, IBM Watsonx.ai, NVIDIA NIM, Hugging Face Smolagents, and Pydantic-AI. The study evaluates these frameworks across several dimensions, such as developer experience, scalability, system performance, and multi-agent collaboration, by analyzing technical documentation and developer resources to understand their architectural capabilities.

\textbf{Qualitative analysis of GitHub discussions}:
Three studies evaluate agentic frameworks by analyzing GitHub developer discussions. For example, Hasan \textit{et al}. \cite{hasan2025empirical} conducted a study analyzing 39 open-source agentic frameworks and 439 agentic applications to investigate common testing practices in the agent ecosystem. Their study examines how developers implement tests within these systems by systematically mining GitHub repositories to extract and categorize test functions. As a result, the authors identify 19 recurring testing patterns, including rule-based, input-driven, and application-level testing.
Similarly, Sun \textit{et al}. \cite{sun2025llm} analyze ten different agentic frameworks, focusing on the datasets used, the underlying LLM architectures, and the roles of agents in the decision-making process, such as communication, planning, and coordination.
Furthermore, Wang \textit{et al}. \cite{wang2025empirical} examine ten agentic frameworks by mining GitHub repositories and developer discussions. Their work investigates widely used frameworks to identify practical challenges in real-world adoption and industrial implementation. The study compares these frameworks across several dimensions, including development efficiency, functional abstraction, learning cost, performance optimization, and maintainability.

\textbf {Summary}:
Despite the rapid growth of agentic frameworks and their increasing adoption in both research and industry, there remains a lack of comprehensive empirical evaluations focusing on reasoning-intensive tasks. As summarized in Table~\ref{tab:agentic_framework_eval}, only two studies have utilized benchmarks to compare a limited number of frameworks, while three studies employ various metrics to evaluate framework performance across different tasks. In contrast, several survey studies have been conducted to compare these frameworks, along with empirical studies that analyze GitHub developer discussions. However, there is a lack of a comprehensive benchmark-based empirical study that systematically selects leading agentic frameworks and evaluates their performance on reasoning tasks using multiple widely adopted benchmarks that encompass a large number of datasets. To address this gap, we have conducted a comprehensive empirical study by systematically selecting widely used agentic frameworks and evaluating their performance on reasoning tasks. In total, we analyzed 22 representative frameworks and assessed their reasoning capabilities across three widely adopted benchmarks. Our evaluation provides insights into the reasoning performance of these frameworks and highlights the trade-offs in terms of accuracy, efficiency, and consistency across different tasks.

\section{Study Design}
\label{Study Design}
\begin{figure}
    \centering
    \includegraphics[width=1.0\linewidth]{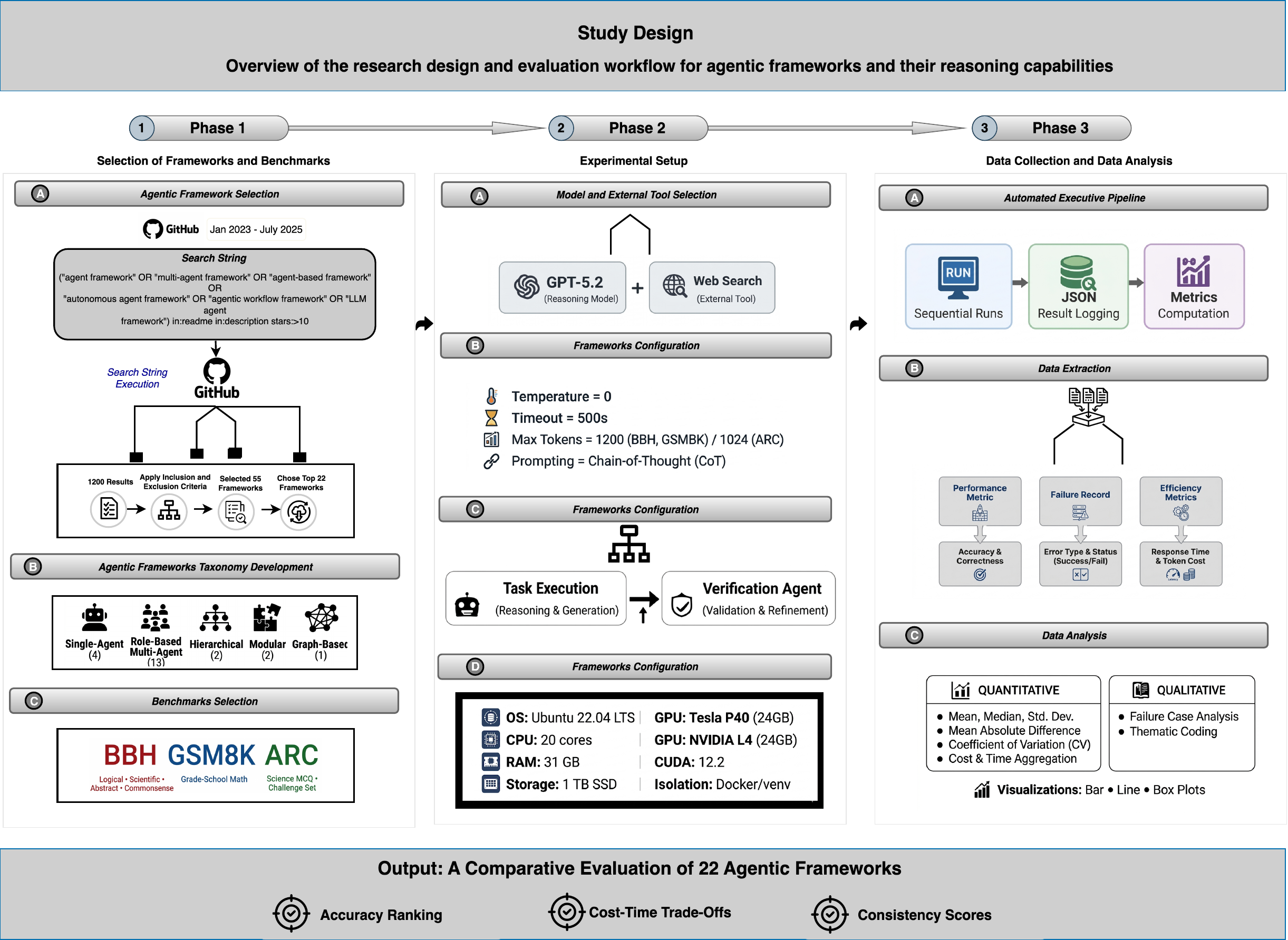}
    \caption{Overview of the research design and evaluation workflow.}
    \label{fig:methodology}
\end{figure}

This study adopts an empirical research methodology using benchmark-based evaluation to assess the reasoning capabilities of agentic frameworks. As shown in Figure~\ref{fig:methodology}, the study is structured around three phases: (i) selecting agentic frameworks and benchmarks, (ii) evaluating the frameworks across multiple benchmarks to measure their reasoning capabilities, and (iii) conducting quantitative and qualitative data analysis. 

\subsection{Research Questions}
\label{Research Questions}
Based on our study goal, we formulated the following three research questions (RQs).

\begin{tcolorbox}[colback=gray!2!white,colframe=black!75!black]
\textit{\textbf{RQ1.} {How effectively do agentic frameworks perform on reasoning tasks across selected benchmarks?}}
\end{tcolorbox}

\textit{\textbf{Objective.}} 
This research question evaluates the effectiveness of multiple agentic frameworks in solving reasoning tasks across selected benchmarks by analyzing both quantitative (e.g., accuracy, success rate) and qualitative (e.g., failure case analysis) performance. The objective is to identify the relative strengths and weaknesses of these frameworks in order to inform future improvements in reasoning methods and benchmark design.

\begin{tcolorbox}[colback=gray!2!white,colframe=black!75!black]
\textit{\textbf{RQ2.} {How do different agentic frameworks compare in terms of cost and time efficiency when performing reasoning tasks across selected benchmarks?}}
\end{tcolorbox}


\textit{\textbf{Objective.}} 
This question investigates the trade-offs between cost and time efficiency across different agentic frameworks when executing reasoning tasks on standardized benchmarks. The objective is to identify the cost- and time-optimal framework for practical deployment.

\begin{tcolorbox}[colback=gray!2!white,colframe=black!75!black]
\textit{\textbf{RQ3.} {To what extent do agentic frameworks demonstrate consistent reasoning performance across the selected benchmarks?}}
\end{tcolorbox}

\textit{\textbf{Objective.}} 
This research question examines the consistency of different agentic frameworks across multiple benchmarks. The objective is to determine whether these frameworks maintain consistent reasoning accuracy across diverse task types, indicating their reliability and generalizability.

\subsection{Agentic Framework Selection}
\label{Agentic Frameworks Selection}

In this study, we collected open-source agentic frameworks from GitHub. The following section describes the selection process and criteria used to identify the frameworks included in our analysis.

\subsubsection{Search String}
Initially, we considered adopting the PICO (Population, Intervention, Comparison, Outcome) framework \cite{schardt2007utilization} to guide the development of our search string. However, applying PICO resulted in an extensive set of keywords and boolean operators, which did not produce relevant results when used as search queries on GitHub. As a result, we developed the final search string based on domain expertise and iterative trial searches conducted collaboratively by the co-authors. This approach allowed us to refine the search strategy in line with the study’s objectives while maintaining a broad scope within practical limits.

In developing our search string, we combined domain-specific terms to identify relevant agentic frameworks. Specifically, we used combinations of the phrases \textit{``agent framework''}, \textit{``multi-agent framework''}, \textit{``agent-based framework''}, \textit{``autonomous agent framework''}, \textit{``agentic workflow framework''}, and \textit{``LLM agent framework''} to ensure broad coverage of repositories describing agent-based or agentic architectures. These terms were searched within both repository \texttt{README} files and description fields (\texttt{in:readme in:description}) to increase the likelihood of retrieving relevant projects that explicitly reference these concepts.

\begin{tcolorbox}[colback=white!2!white,colframe=white!75!black]
\textit{\textbf{("agent framework" OR "multi-agent framework" OR "agent-based framework" OR "autonomous agent framework" OR "agentic workflow framework" OR "LLM agent framework") in:readme in:description stars:$>10$}}
\end{tcolorbox}

The initial search produced 1,200 repositories as of 26 July 2025. In the next phase, a manual analysis of the retrieved repositories was conducted. Based on the predefined inclusion and exclusion criteria presented in Table~\ref{tab:inc_exc_criteria}, a final set of agentic frameworks was selected.


\begin{table}[t]
\centering
\renewcommand{\arraystretch}{1.2}
\begin{tabularx}{\columnwidth}{|l|l|X|}
\hline
\textbf{ID} & \textbf{Type} & \textbf{Criteria Description} \\
\hline
I1 & Inclusion & The repository must have more than 10 stars and 10 forks. \\
\hline
I2 & Inclusion & It must contain at least one open issue. \\
\hline
I3 & Inclusion & The project should have more than 5 contributors. \\
\hline
I4 & Inclusion & The project must genuinely be a framework for agent-based systems, not just contain the keywords. \\
\hline
I5 & Inclusion & Project documentation and content must be in English. \\
\hline
E1 & Exclusion & The repository is still under development. \\
\hline
E2 & Exclusion & The repository is set to read-only mode. \\
\hline
E3 & Exclusion & The repository is no longer active or maintained. \\
\hline
E4 & Exclusion & The repository content is not in English. \\
\hline
\end{tabularx}
\caption{Inclusion and exclusion criteria for agentic framework selection}
\label{tab:inc_exc_criteria}
\end{table}

\subsubsection{Framework Selection}

Initially, we manually analyzed a total of 1,200 repositories retrieved from GitHub to identify potential agentic frameworks. In the first phase, based on the inclusion and exclusion criteria described in Table~\ref{tab:inc_exc_criteria}, we shortlisted 55 repositories that met the minimum quality and relevance requirements. More details about the frameworks can be found in \cite{rasheed_2026_19593242}.

\subsubsection{Finalized Frameworks}
For further experimentation, we selected 22 agentic frameworks from the 55 shortlisted frameworks based on the following criteria:

\begin{itemize}
\item The number of GitHub stars and forks, used to select the top 22 frameworks with the most number of GitHub stars and forks.

\item The number of contributors and recent commits, used to select the top 22 frameworks with the most active development.

\item A healthy number of open and closed issues.

\end{itemize}

Applying these criteria resulted in a final set of 22 agentic frameworks for empirical analysis and evaluation. The selected frameworks are presented in Table~\ref{tab:agentic_taxonomy}.

\subsection{Classification of Agentic Frameworks}

To analyze the agentic frameworks systematically, we classified the 22 selected frameworks into five categories based on their primary architectural focus and core functionality: (i) single-agent architecture, (ii) role-based multi-agent architecture, (iii) hierarchical architecture, (iv) modular component architecture, and (v) graph-based workflow frameworks. To categorize the selected frameworks into these five categories, we followed previous studies such as \cite{qiao2023taskweaver}, \cite{vaidhyanathan2025agentic}, and \cite{patel2025systematic}.

\subsubsection{Single-Agent Architecture}

Single-agent architectures focus on planning, observation, and action within the capabilities of a single autonomous agent \cite{qiao2023taskweaver}. As illustrated in Table \ref{tab:agentic_taxonomy}, four frameworks fall into this category: AutoGPT, TaskWeaver, Semantic-kernel, and LangChain.

\subsubsection{Role-Based Multi-Agent Architecture}

Role-based multi-agent architecture frameworks focus on collaboration among multiple autonomous agents that interact through structured, role-based communication, often combining predefined coordination rules with LLM-driven decision-making \cite{derouiche2025agentic}. As shown in Table \ref{tab:agentic_taxonomy}, 13 frameworks were classified under this category, i.e., BabyAGI, AutoGen, Camel, CrewAI, SuperAGI, Swarm, Agency-Swarm, OpenAI-Agents-Python, Agent-zero, PraisonAI, Qwen-Agent, Pydantic-AI, and ANUS. These frameworks are designed around agent-to-agent communication and collaboration as the primary mechanisms for task execution.

\subsubsection{Hierarchical Architecture}
Hierarchical architectures organize agents in a multi-level structure where higher-level agents are responsible for planning, coordination, and task decomposition, while lower-level agents execute specific subtasks \cite{vaidhyanathan2025agentic}. As shown in Table \ref{tab:agentic_taxonomy}, two frameworks are classified in this category: MetaGPT and Google ADK. These frameworks highlight structured task management, where responsibilities are distributed across different layers of agents to improve coordination and scalability.

\subsubsection{Modular Component Architecture}
Modular component architectures are designed around the integration of specialized and reusable components that can be combined to extend agent capabilities and facilitate interaction with external systems \cite{patel2025systematic}. In our taxonomy, two frameworks fall under this category: Mastra and Upsonic. These frameworks highlight the importance of modularity, allowing developers to build flexible agent systems by assembling independent functional components.

\subsubsection{Graph-Based Workflow Frameworks}
Graph-based workflow frameworks represent the execution and coordination of agents as directed graphs, where nodes correspond to agents or tasks and edges capture control flow, state transitions, and dependencies \cite{wang2024agent}, \cite{patel2025systematic}. LangGraph is classified within this category due to its explicit graph-based execution model. As noted by Derouiche \textit{et al}.~\cite{derouiche2025agentic}, LangGraph introduces a novel graph-based approach for sequencing tasks among LLM agents. By supporting compositional workflows and stateful operations, it enables traceable and scalable agent design, particularly in research and analytics contexts.

\begin{table*}[t]
\centering
\renewcommand{\arraystretch}{1.2}
\small
\resizebox{\textwidth}{!}{
\begin{tabular}{|c|c|c|c|c|c|L{8.8cm}|}
\hline
\textbf{Primary Category} & \textbf{Agentic Framework} & \textbf{Reasoning} & \textbf{GitHub Stars} & \textbf{Forks} & \textbf{Company Name} & \textbf{GitHub Link} \\
\hline

\multirow{4}{*}{Single-Agent Architecture}
& AutoGPT & CoT & 182k & 46.2k & Significant-Gravitas & https://github.com/Significant-Gravitas/AutoGPT \\
& TaskWeaver & TF & 6.1k & 759 & Microsoft & https://github.com/microsoft/TaskWeaver \\
& Semantic-kernel & TF & 27.4k & 4.5k & Microsoft & https://github.com/microsoft/semantic-kernel \\
& LangChain & CoT & 110k & 21.1k & langchain-ai & https://github.com/langchain-ai/langchain \\
\hline

\multirow{13}{*}{Role-Based Multi-Agent Architecture}
& BabyAGI & CoT & 22.2k & 2.9k & yoheinakajima & https://github.com/yoheinakajima/babyagi \\
& Autogen & Hybrid & 55.3k & 8.3k & Microsoft & https://github.com/microsoft/autogen \\
& Camel & CoT & 16.2k & 1.8k & camel-ai & https://github.com/camel-ai/camel \\
& CrewAI & Hybrid & 45.5k & 6.1k & crewAIInc & https://github.com/crewAIInc/crewAI \\
& SuperAGI & Hybrid & 17.2k & 2.2k & TransformerOptimus & https://github.com/TransformerOptimus/SuperAGI \\
& Swarm & Hybrid & 21.1k & 2.2k & OpenAI & https://github.com/openai/swarm \\
& Agency-swarm & Hybrid & 4k & 1k & VRSEN & https://github.com/VRSEN/agency-swarm \\
& OpenAI-Agents-Python & Hybrid & 19.4k & 3.2k & OpenAI & https://github.com/openai/openai-agents-python \\
& Agent-zero & CoT & 15.9k & 3.3k & agent0ai & https://github.com/agent0ai/agent-zero \\
& PraisonAI & Hybrid & 5.6k & 773 & MervinPraison & https://github.com/MervinPraison/PraisonAI \\
& Qwen-Agent & Hybrid & 15.2k & 1.5k & QwenLM & https://github.com/QwenLM/Qwen-Agent \\
& Pydantic-AI & TF & 15.3k & 1.7k & pydantic & https://github.com/pydantic/pydantic-ai \\
& ANUS & Hybrid & 6.3k & 921 & anus-dev & https://github.com/anus-dev/ANUS \\
\hline

\multirow{2}{*}{Hierarchical Architecture}
& MetaGPT & Hybrid & 64.9k & 8.2k & FoundationAgents & https://github.com/FoundationAgents/MetaGPT \\
& Google ADK & Hybrid & 18.2k & 3k & Google & https://github.com/google/adk-python \\
\hline

\multirow{2}{*}{Modular Component Architecture}
& Upsonic & TF & 7.8k & 722 & Upsonic & https://github.com/Upsonic/Upsonic \\
& Mastra & TF & 21.8k & 1.7k & mastra-ai & https://github.com/mastra-ai/mastra \\
\hline

Graph-Based Architecture & LangGraph & Hybrid & 25.9k & 4.5k & langchain-ai & https://github.com/langchain-ai/langgraph \\
\hline
\end{tabular}
}
\caption{Taxonomy of selected agentic frameworks based on architectural design.}
\label{tab:agentic_taxonomy}
\end{table*}


\subsection{Benchmark Selection}
\label{Benchmarks Selections}

We searched for existing benchmarks that satisfy the following specific selection criteria: (i) The benchmarks should be well-established within the research community and frequently used for evaluating reasoning performance, ensuring that our results are comparable and credible. (ii) The selected benchmarks should include a diverse set of reasoning tasks that cover multiple reasoning dimensions, such as logical, scientific, mathematical, commonsense, and abstract reasoning \cite{huang2023towards}. (iii) Each benchmark should provide publicly available datasets and evaluation protocols to facilitate reproducibility and transparency. (iv) The benchmark dataset must be suitable for agentic AI evaluation.

Initially, we searched for suitable benchmarks based on the study conducted by Ferrag \textit{et al}. \cite{ferrag2025llm}. This study presents a comprehensive comparison of benchmarks developed between 2018 and 2025 that evaluate language models and agents across multiple domains. Based on the selection criteria described above, we chose the BBH benchmark~\cite{suzgun2022challenging}, which is widely recognized for its emphasis on multi-step and abstract reasoning \cite{kazemi2025big}. BBH comprises 27 distinct reasoning tasks, including Boolean logic, causal judgment, arithmetic reasoning, and other forms of analytical and deductive reasoning. 

To evaluate mathematical reasoning, we selected the GSM8K benchmark~\cite{cobbe2021training}, which includes grade-school-level mathematical word problems. The dataset specifically focuses on word problems that involve real-world contexts such as calculating costs, determining quantities, working with rates and ratios, and solving problems involving time, distance, and other practical mathematical applications \cite{zhang2022automatic}.
Additionally, to assess scientific reasoning, we utilized the ARC benchmark~\cite{clark2018think}, which consists of multiple-choice science questions designed to test factual understanding and reasoning ability. This benchmark contains a comprehensive dataset containing grade-school-level science questions sourced from actual human standardized tests. The benchmark is divided into two partitions: Easy and Challenge, where the challenge set specifically contains questions that cannot be answered through simple information retrieval or statistical word co-occurrence methods, thus requiring genuine reasoning capabilities  \cite{clark2018think}.

Table~\ref{tab:benchmarks} presents an overview of the three selected benchmarks, highlighting their reasoning task types and the number of tasks included in each.

\begin{table}[htbp]
\centering
\resizebox{\textwidth}{!}{
\begin{tabular}{|c|l|l|c|l|}
\hline
\textbf{S. No.} & \textbf{Benchmark}            & \textbf{Reasoning Task Type}                                & \textbf{Paper Reference}                                                              \\ \hline
1               & BBH          & 27 challenges covering various reasoning types    & Suzgun \textit{et al}. \cite{suzgun2022challenging} \\ \hline
2               & GSM8K                         & Mathematical reasoning (grade-school-level problem solving) & Cobbe \textit{et al}. \cite{cobbe2021training}   \\ \hline
3               & ARC & Scientific reasoning (multiple-choice science questions)    & Clark \textit{et al}. \cite{clark2018think}        \\ \hline
\end{tabular}
}
\caption{Overview of selected benchmarks and their reasoning tasks.}
\label{tab:benchmarks}
\end{table}

\subsection{Experiment Setup}
\label{Experiment Setup}

In this section, we present the experimental details of the agentic framework development setup. We describe the models and external tools used for agent development and execution. We also detail the configuration, environment, and evaluation setup of the selected agentic frameworks, as well as the hardware platforms used in this project.

\subsubsection{Language Model}

In this study, we utilized the GPT-5.2 model for all experiments. The same model was used across all agentic frameworks to maintain experimental control and ensure fair and valid framework comparisons. 


We selected GPT-5.2 as the base model for all agentic framework evaluations because of its strong reasoning capabilities and high performance on standardized benchmarks~\cite{openai_gpt52_2025}. On the ARC-AGI-1~\cite{liao2025arc} and ARC-AGI-2~\cite{chollet2025arc} benchmarks, GPT-5.2 achieves accuracies of 86.2\% and 52.1\%, respectively, outperforming Gemini 3 Pro and Opus 4.5 \cite{arcprize_leaderboard}.
Similarly, on the CharXiv Reasoning benchmark~\cite{wang2024charxiv}, which focuses on complex reasoning tasks requiring information synthesis, GPT-5.2 achieves a leading score of 0.821, compared to 0.814 by Gemini 3 Pro. On FrontierMath (Tier 1–3)~\cite{glazer2024frontiermath}, a benchmark designed to evaluate advanced mathematical reasoning, GPT-5.2 achieves the highest accuracy of 40.3\% among the models evaluated.

Collectively, these benchmarks address abstract reasoning, advanced mathematical reasoning, and scientific problem-solving tasks. In addition, GPT-5.2 demonstrates strong performance on other established benchmarks, including SWE-Bench (Verified), GPQA Diamond, and AIME, further supporting its suitability as a robust base model for comparing the reasoning capabilities of different agentic frameworks~\cite{openai_gpt52_2025}.

\subsubsection{External Environment}

In this experiment, we used the GPT-provided external \textit{web search} \footnote{\url{https://platform.openai.com/docs/guides/tools-web-search}} tool to enable access to information beyond the model’s internal training data. The \textit{web search} tool allows agents to query online sources when the required information cannot be obtained from the language model’s internal knowledge. This tool-augmented setup enables agents to retrieve up-to-date and task-specific information at runtime, which is particularly beneficial for knowledge-intensive and dynamically evolving tasks. By integrating \textit{web search} while preserving the LLM as the core reasoning component, agentic frameworks can enhance response accuracy and contextual awareness without altering the underlying decision-making process of the model.

\subsubsection{Framework Configuration}
Each agentic framework was configured under a standardized setup to ensure comparability and reproducibility across the evaluations. All frameworks were initialized within a unified execution environment that maintained identical model parameters, including a fixed few-shot configuration and chain-of-thought prompting for step-by-step reasoning. The temperature parameter was set to 0 to eliminate randomness~\cite{zhang2022automatic}.
The token limit for each benchmark was determined through a pilot experiment. Initially, we adopted token limits suggested by prior studies. For example, previous work reports an average limit of 224 tokens per task for BBH~\cite{zhou2024self}, while Zhang \textit{et al}.~\cite{zhang2022automatic} used a limit of 256 tokens for GSM8K to support step-by-step mathematical reasoning. For ARC, we initially selected a token limit of 224 tokens because the benchmark consists of multiple-choice questions with relatively short expected answers. Accordingly, we first applied these token limits in our pilot experiments. However, based on the initial results, we observed that these limits were insufficient for our multi-agent setup, as agents frequently required additional tokens to generate complete reasoning steps and final answers when using chain-of-thought reasoning. The previous studies used smaller token limits because their experiments did not involve multi-agent communication and extended reasoning steps. Therefore, based on the pilot results, we increased the token limits to avoid cutting off responses and to allow agents to produce full reasoning outputs. As a result, we set the maximum token limit to 1200 tokens for BBH and GSM8K.
For the ARC benchmark, our pilot experiments indicated that a smaller limit was sufficient due to the shorter answer format. Therefore, we set the maximum token limit to 1024 tokens for ARC, which allows sufficient reasoning while keeping the responses concise. In addition, we set the API timeout to 500 seconds to ensure consistent response behavior across all benchmarks.

\subsubsection{Environment Setup and Dependencies}
All agentic frameworks were deployed within isolated virtual environments to ensure dependency integrity and consistency throughout the experimental process. Each environment was initialized using a unified dependency management system that synchronized library versions and Python configurations across all frameworks. Core dependencies, including libraries for model interaction, data processing, and evaluation management, were version-locked to ensure consistent behavior and stable execution across the experiments.

\subsubsection{Multi-Agent Setup}
To evaluate the reasoning capabilities of different agentic frameworks, we designed a multi-agent setup and applied it consistently across 22 agentic frameworks. For each framework, we developed agents that execute reasoning tasks defined by three standardized benchmarks. In this project, we adopted a two-agent setup. We conducted a pilot study to experiment with different numbers of agents. For instance, a single-agent setup does not allow evaluation of agent communication, coordination, and other core characteristics of agentic frameworks that cannot be examined in single-agent settings. On the other hand, configurations with a larger number of agents (e.g., four agents) significantly increased the execution time and API cost due to more frequent interactions between agents. Finally, based on the results of the pilot experiments, we selected a two-agent setup as a balanced configuration that enables the evaluation of agent interaction while keeping the computational cost and execution time manageable.

In this setup, the \textit{Task Execution Agent} is responsible for performing benchmark-specific reasoning tasks. It processes each input question from the benchmark dataset, systematically applies chain-of-thought prompting techniques to generate step-by-step reasoning, and produces candidate solutions. These outputs are then forwarded to the \textit{Verification Agent} for consistency checking and refinement. The \textit{Verification Agent} independently evaluates the outputs produced by the \textit{Task Execution Agent}, by checking the correctness, consistency, and reasoning quality of the generated answers. To ensure consistent and controlled comparisons across frameworks, we have kept the agent roles, prompts, task descriptions, and verification criteria the same for all frameworks. Fixing the multi-agent configuration and prompt design ensures that performance differences can be attributed to the underlying agentic frameworks rather than variations in agent behavior or prompt engineering.

\subsubsection{Hardware Platforms}
Experiments were conducted on a server running Debian GNU/Linux 12 (Bookworm). The system is equipped with an Intel Core i7-14700 processor (20 cores, 28 threads), 31 GB RAM, and a 952 GB NVMe SSD. The server includes two GPUs: an NVIDIA Tesla P40 with 24 GB VRAM and an NVIDIA L4 with 24 GB VRAM. The system uses NVIDIA driver version 535.216.01 with CUDA 12.2 support.

\subsubsection{Evaluation Setup}

For the evaluation phase, each benchmark was executed through an automated evaluation pipeline that sequentially loaded the benchmark dataset, set up the corresponding agentic framework, and executed the reasoning tasks. Tasks within each benchmark were processed sequentially to ensure consistent execution and resource usage across frameworks. 
Model inference was performed with a predefined retry policy to handle transient failures, such as API timeouts or rate-limit errors. Failed requests were retried up to a fixed number of attempts before being marked as unsuccessful and logged accordingly. Tasks exceeding a predefined execution time limit were terminated and recorded as timeouts.

The system supported both sample-mode evaluations, using a small subset of tasks for rapid testing and debugging, and full-run evaluations, covering the complete benchmark dataset for comprehensive analysis. All generated responses were validated against benchmark-specific answer formats and automatically scored using the official evaluation criteria provided by each benchmark.
During execution, the pipeline logged detailed metadata for each task, including task identifiers, execution time, number of retries, success or failure status, and final scores. Performance metrics such as accuracy, task completion rate, and total runtime were computed and stored for each framework and benchmark combination. This evaluation setup enabled systematic and reproducible comparison of agentic frameworks across multiple reasoning benchmarks.

\subsection{Data Collection}

In this study, experimental results were systematically collected for each evaluated agentic framework across all selected benchmarks. All experiments were executed in a standardized and controlled computational environment. This controlled setup ensures that the collected data accurately reflect framework-level behavior rather than external infrastructure effects.

For each framework–benchmark execution, the generated outputs and associated metadata were automatically recorded and stored. The collected data include task results, accuracy scores, framework response times, computational cost, and other evaluation-related statistics. All results were saved in a structured JSON format to support reproducibility and downstream analysis. To maintain clear organization and traceability, results were stored separately for each benchmark and framework (e.g., FrameworkName\_BBH\_config.log.json, FrameworkName \_GSM8K\_config.log..json, FrameworkName\_ARC\_config.log.json), enabling modular analysis and facilitating comparison across benchmarks during the evaluation phase.

\subsection{Data Analysis}

In this study, we analyzed multiple types of data using both quantitative and qualitative approaches. For the quantitative analysis, descriptive statistical methods \cite{wohlin2006empirical} were employed. The collected results were examined using a combination of descriptive statistics (e.g., mean, median, standard deviation, standard error of the mean (SEM), F-statistics, P-value,  mean absolute difference, and coefficient of variation (CV)). For the qualitative data, we utilized thematic analysis \cite{blair2015reflexive} to analyze open-ended responses. To facilitate clearer interpretation of the experimental outcomes, various data visualizations—including bar charts, line graphs, and box plots—were used for effective presentation and comparison of the results. In the following sections, we discuss the data analysis methods employed for each research question (RQ) in detail.

\subsubsection{Accuracy (RQ1)}

To address RQ1, we evaluated the reasoning performance of the agentic frameworks using three widely adopted reasoning benchmarks: BBH, GSM8K, and ARC. We first applied descriptive statistical analysis to compute the accuracy of each framework on individual benchmarks. To summarize the overall performance of each framework, the benchmark-level accuracy scores were combined by calculating their mean across the three benchmarks. This process produces a single representative accuracy score for each framework, enabling a clear and holistic comparison of reasoning performance across diverse reasoning tasks. Finally, the frameworks were ranked according to their mean accuracy scores to facilitate comparative analysis of their reasoning capabilities.

\[
\text{Mean Accuracy} = \bar{x} = \frac{1}{n} \sum_{i=1}^{n} x_i
\]

\textbf{Framework-level performance variation analysis}:
To understand how the performance of each framework varies across the evaluated benchmarks, the SEM is reported as an additional measure in the performance analysis. SEM reflects the extent to which the benchmark accuracies of a framework vary around its mean accuracy. In this study, a smaller SEM indicates that the benchmark results are relatively close to one another, whereas a larger SEM reflects greater differences among the benchmark accuracies.

Since the SEM is calculated from the standard deviation, the standard deviation is first defined as

\begin{equation}
SD = \sqrt{\frac{\sum_{i=1}^{n}(x_i-\bar{x})^2}{n-1}},
\end{equation}

where $x_i$ denotes the accuracy value of a framework on benchmark $i$, $\bar{x}$ denotes the mean accuracy of that framework across the available benchmarks, and $n$ denotes the number of available benchmark results.

The SEM is then computed as

\begin{equation}
SEM = \frac{SD}{\sqrt{n}},
\end{equation}

where $SD$ denotes the standard deviation of the benchmark accuracies for a given framework and $n$ denotes the number of available benchmark results used to compute the mean. In this study, $n = 3$ for frameworks with valid results on all three benchmarks, while smaller values of $n$ are used for frameworks with incomplete benchmark results.

\textbf{Architecture-based performance analysis:}
To understand how framework performance varies within each architectural category, we employed one-way analysis of variance (ANOVA). This method was used to test whether the benchmark accuracies of frameworks in the same category differ significantly across the three evaluated benchmarks. ANOVA is appropriate in this context because it enables the comparison of mean performance across multiple benchmark groups within a single statistical framework.

The ANOVA test is based on the F-statistic, which is calculated as

\begin{equation}
F = \frac{MS_{\text{between}}}{MS_{\text{within}}},
\end{equation}

where $MS_{\text{between}}$ denotes the mean square between groups, representing the variation among benchmark means, and $MS_{\text{within}}$ denotes the mean square within groups, representing the variation within each benchmark group. A larger $F$ value indicates that the variation between benchmark means is greater relative to the variation within groups.

To determine whether the observed variation is statistically significant, the corresponding p-value was also computed. In this study, statistical significance was determined using a threshold of $p < 0.05$. A p-value below this threshold indicates that the observed performance differences across benchmarks are unlikely to have occurred by chance, whereas a p-value above this threshold suggests that the variation is not statistically significant.

\textbf{Failure modes:}
We also analyzed the data to identify failure modes in which the frameworks show differing behaviors. We applied thematic analysis to qualitatively analyze and characterize the observed failure patterns.

\subsubsection{Cost+Time (RQ2)}

To evaluate the computational \textbf{cost} of each agentic framework, we recorded the total number of tokens consumed for each benchmark task. The overall token consumption was then calculated by summing the tokens used across all tasks, as defined in Eq.~(\ref{eq1}):

\begin{equation}
\text{Total Tokens} = \sum_{i=1}^{N} T_i
\label{eq1}
\end{equation}

To obtain a normalized measure of computational cost, we computed the average token consumption per task using the arithmetic mean, as shown in Eq.~(\ref{eq2}):

\begin{equation}
\text{Average Tokens per Task} = \frac{1}{N} \sum_{i=1}^{N} T_i
\label{eq2}
\end{equation}

Here, $T_i$ represents the number of tokens consumed for the $i$-th task, and $N$ denotes the total number of benchmark tasks.

Finally, the cost associated with each framework was calculated based on the total and average token consumption. Specifically, both the total cost and the average cost per task were estimated using the current OpenAI token pricing. As the GPT-5.2 model was used in this study, the cost calculation was performed according to its pricing at the time of experimentation.

\textbf{Time:}
To evaluate the \textbf{time efficiency} of the agentic frameworks, we measured the execution time required to complete each benchmark task. For every task, the total response time was recorded as the elapsed wall-clock time between the start of execution and the generation of the final output. The overall time consumption was then calculated by summing the execution time across all tasks. In addition, the average time per task was computed using the arithmetic mean to provide a normalized measure of performance.

\subsubsection{Consistency Analysis (RQ3)}
To address RQ3, we evaluated the \textbf{consistency} of agentic frameworks across multiple benchmarks by computing both absolute and relative (scale-adjusted) consistency measures. Specifically, we analyzed the consistency of each framework across the three benchmarks (BBH-ARC, BBH-GSM8K, and ARC-GSM8K).

To measure absolute consistency, we employed an absolute difference-based approach. This method computes the pairwise absolute differences between benchmark accuracies and aggregates them to obtain a total deviation score for each framework. A lower total deviation indicates higher consistency across benchmarks. The total deviation is calculated as shown in Eq.~(\ref{eq3}):

\begin{equation}
\text{Total Deviation} = |A_1 - A_2| + |A_1 - A_3| + |A_2 - A_3|
\label{eq3}
\end{equation}

To provide a normalized measure of variation, we further computed the mean deviation across benchmark pairs, as defined in Eq.~(\ref{eq4}):

\begin{equation}
\text{Mean Deviation} = \frac{|A_1 - A_2| + |A_1 - A_3| + |A_2 - A_3|}{3}
\label{eq4}
\end{equation}

To measure relative (scale-adjusted) consistency, we utilized the CV method, which quantifies variability relative to the mean performance across benchmarks. The CV is defined in Eq.~(\ref{eq5}):

\begin{equation}
\text{CV} = \frac{\sigma}{\mu},
\label{eq5}
\end{equation}

where $\sigma$ represents the standard deviation and $\mu$ denotes the mean accuracy across the three benchmarks. In this study, a lower CV value indicates higher consistency in framework performance.

\section{Experimental Results}
\label{Result}
In this section, we present the outcome of our benchmark-based evaluation of 22 agentic frameworks. As shown in Table \ref{tab:agentic_benchmark}, the agentic frameworks are divided into five categories based on their architecture. The results section is divided according to the three research questions presented in Section \ref{Research Questions}.

\subsection{Agentic Framework Performance (RQ1)}
\label{Agentic Framework Performance}

\begin{figure}
    \centering
    \includegraphics[width=0.9\linewidth]{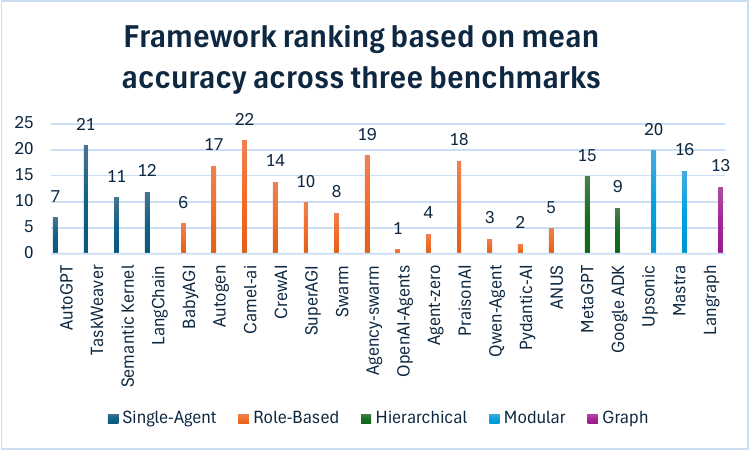}
    \caption{Comparative ranking of agentic frameworks based on the mean accuracy across three benchmarks (Rank 1 is the highest).}
    \label{fig:rank}
\end{figure}

We evaluated the performance of agentic frameworks on reasoning tasks using three benchmarks, namely BBH, GSM8K, and ARC. 
Table \ref{tab:agentic_benchmark} reports the accuracy of each agentic framework on the three benchmarks, together with its mean score, SEM, and completion. The mean score serves as an overall basis for comparing framework performance across the benchmarks. A more detailed comparison based on these mean scores is presented below and further illustrated in Figure~\ref{fig:rank}.
In addition to mean accuracy, the SEM is reported to reflect the variation in each framework’s accuracy across the benchmarks. A smaller SEM indicates that the accuracy levels of the framework across the three benchmarks are relatively close to each other, whereas a larger SEM reflects greater differences in framework performance across benchmarks. For example, TaskWeaver and Agency-Swarm show relatively low SEM values of 3.12 and 2.98, respectively, indicating limited variation across benchmarks, although their accuracy remains consistently low. In contrast, frameworks such as Mastra, MetaGPT, AutoGen, and PraisonAI exhibit high SEM values of 26.55, 27.93, 26.16, and 25.08, respectively, owing to substantial fluctuations in performance across the benchmarks. Overall, most frameworks show SEM values ranging approximately from 12 to 16. We have also reported completion to indicate the number of benchmarks for which valid results were obtained for each framework. Most frameworks completed all three benchmarks, whereas Camel completed only one benchmark, and Upsonic and Mastra completed two.

To further illustrate comparative performance, Figure~\ref{fig:rank} presents the ranking of the agentic frameworks based on their overall mean scores across the three benchmarks.
Overall, 12 frameworks achieved mean accuracies ranging from 74.57\% to 75.94\%, with only a 1.4 percentage point difference between the lowest and highest values. Among these frameworks, those in the role-based multi-agent category tended to achieve slightly higher accuracy. For example, OpenAI-Agents-Python, Pydantic-AI, Qwen-Agent, Agent-zero, ANUS, and BabyAGI ranked first, second, third, fourth, fifth, and sixth, respectively. However, some well-known frameworks in this category, such as AutoGen and Camel, showed poorer performance, ranking 18th and 22nd, respectively. The reasons for the failure of these frameworks are provided in Section~\ref{Observed Failures}.

\begin{table}[t]
\centering
\resizebox{\textwidth}{!}{
\begin{tabular}{|l|l|c|c|c|c|c|c|}
\hline
\textbf{Primary Category} & \textbf{Agentic Framework} & \textbf{BBH (\%)} & \textbf{GSM8K (\%)} & \textbf{ARC (\%)} & \textbf{Mean (\%)} & \textbf{SEM} & \textbf{Completion} \\ \hline

\multirow{4}{*}{Single-Agent Architecture}
& AutoGPT & 90.77 & 44.40 & 90.76 & 75.31 & 15.46 & 3/3 \\ 
& TaskWeaver & 13.88 & 7.92 & 18.70 & 13.50 & 3.12 & 3/3 \\ 
& Semantic-kernel & 89.16 & 44.49 & 90.94 & 74.86 & 15.20 & 3/3 \\ 
& LangChain & 89.86 & 44.39 & 90.05 & 74.77 & 15.19 & 3/3 \\ \hline

\multirow{13}{*}{Role-Based Multi-Agent Architecture}
& BabyAGI & 89.77 & 44.48 & 91.78 & 75.34 & 15.44 & 3/3 \\ 
& AutoGen & 24.08 & 4.55 & 90.94 & 39.86 & 26.16 & 3/3 \\ 
& Camel & -- & -- & 8.75 & -- & -- & 1/3 \\ 
& CrewAI & 78.91 & 42.82 & 91.51 & 71.08 & 14.59 & 3/3 \\ 
& SuperAGI & 90.11 & 44.51 & 90.49 & 75.04 & 15.26 & 3/3 \\ 
& Swarm & 90.43 & 44.29 & 90.72 & 75.15 & 15.43 & 3/3 \\ 
& Agency-Swarm & 34.57 & 37.58 & 27.51 & 33.22 & 2.98 & 3/3 \\ 
& OpenAI-Agents-Python & 91.06 & 44.31 & 92.45 & 75.94 & 15.82 & 3/3 \\ 
& Agent-zero & 89.48 & 44.21 & 92.58 & 75.42 & 15.63 & 3/3 \\ 
& PraisonAI & 20.30 & 4.66 & 86.49 & 37.15 & 25.08 & 3/3 \\ 
& Qwen-Agent & 90.69 & 44.33 & 92.54 & 75.85 & 15.77 & 3/3 \\ 
& Pydantic-AI & 90.71 & 44.44 & 92.67 & 75.94 & 15.76 & 3/3 \\ 
& ANUS & 89.83 & 44.21 & 92.18 & 75.41 & 15.61 & 3/3 \\ \hline

\multirow{2}{*}{Hierarchical Architecture}
& MetaGPT & 72.75 & 0.00 & 91.60 & 54.78 & 27.93 & 3/3 \\ 
& Google ADK & 90.49 & 44.57 & 90.15 & 75.07 & 15.25 & 3/3 \\ \hline

\multirow{2}{*}{Modular Component Architecture}
& Upsonic & -- & 43.29 & 67.61 & 23.63 & 22.01 & 2/3 \\ 
& Mastra & -- & 32.46 & 90.77 & 41.08 & 26.55 & 2/3 \\ \hline

Graph-Based Architecture & LangGraph & 89.80 & 44.35 & 89.56 & 74.57 & 15.11 & 3/3 \\ \hline
\end{tabular}
}
\caption{Accuracy of agentic frameworks across three benchmarks, mean accuracy (higher percentages indicate better results), SEM (lower values indicate less variation), and completion (valid benchmark results).}
\label{tab:agentic_benchmark}
\end{table}

The frameworks in the single-agent architecture category exhibit notable variation in performance. Of these frameworks, AutoGPT achieved consistently high accuracy across the evaluated benchmarks, securing an overall rank of seventh. Similarly, Semantic-kernel and LangChain achieved moderate accuracy, ranking 11th and 12th, respectively. In contrast, TaskWeaver ranked 21st overall, reflecting lower average accuracy compared to other frameworks. Overall, the results indicate that, while certain single-agent frameworks can maintain competitive performance, others may struggle to generalize effectively across diverse benchmark tasks.

Similarly, within the hierarchical architecture category, performance varies across frameworks. For example, MetaGPT ranked 15th overall, while Google ADK ranked 9th, indicating comparatively stronger performance across the evaluated benchmarks.

Within the modular component architecture category, both frameworks show relatively lower rankings. For example, the Upsonic framework ranked 20th and also exhibited the highest token consumption during execution. This high computational cost, combined with relatively weak performance, highlights potential inefficiencies in task orchestration and reasoning control mechanisms. A detailed analysis of execution failures, runtime behavior, and cost-related challenges is provided in Section~\ref{Observed Failures}. 
Similarly, the Mastra framework ranked 16th overall, indicating lower mean performance.

In the graph-based architecture category, LangGraph ranked 13th, indicating moderate performance compared to the other evaluated frameworks. More details about the dataset can be found in our replication package \cite{rasheed_2026_19593242}.

\subsubsection{Performance Variation Across Categories}
As shown in Table~\ref{tab:anova_new}, the statistical significance of performance differences across frameworks within each category was evaluated using the F-statistic and $p$-value approach. The results indicate that the role-based multi-agent architecture category exhibits a statistically significant variance \((F(2, 33) = 16.61,\, p = 1.02 \times 10^{-5})\).
This highlights the fact that framework performance within this category varies considerably, suggesting that specific architectural designs and task characteristics strongly influence their effectiveness. For instance, although the highest-ranking frameworks belong to this category, it also includes lower-performing frameworks, resulting in significant differences in overall performance.

In contrast, the single-agent architecture and hierarchical architecture categories do not show statistically significant differences across the frameworks, as indicated by their respective $p$-values of $0.231$ and $0.071$. This suggests a higher degree of performance consistency among frameworks within these groups. However, the limited number of frameworks in some categories, particularly in the hierarchical architecture category, may reduce the statistical power and affect the reliability of these comparisons.

\begin{table}[h]
\centering
\begin{tabular}{|lccc|}
\hline
\textbf{Category} & \textbf{F-Statistic} & \textbf{p-value} & \textbf{Significant} \\
\hline
Single-Agent Architecture & $F(2,9) = 1.73$ & 0.231 & No \\
Role-Based Multi-Agent Architecture & $F(2,33) = 16.61$ & $1.02\times10^{-5}$ & Yes \\
Hierarchical Architecture & $F(2,3) = 7.22$ & 0.071 & No \\
Modular Component Architecture & -- & -- & No \\
Graph-Based Architecture & -- & -- & No \\
\hline
\end{tabular}
\caption{ANOVA results for performance variance across agentic framework categories.}
\label{tab:anova_new}
\end{table}

For the modular component architecture and graph-based architecture categories, statistical comparison was not feasible due to insufficient benchmark results or the presence of only a single framework.

Overall, the findings suggest that, while role-based multi-agent architectures offer the highest performance potential, they also show significant internal variance, indicating that architectural complexity does not guarantee superior results. These results suggest that for complex reasoning and mathematical tasks, the choice of a specific multi-agent design is more effective than the broad selection of a category, as individual framework maturity remains a key factor in overall performance.

\subsubsection{Observed Failures and Practical Constraints}
\label{Observed Failures}

The Camel framework demonstrated limited performance across the evaluated benchmarks. It was only able to fully execute the ARC benchmark, where it achieved an accuracy of 8.75\%. However, for the BBH and GSM8K benchmarks, Camel failed to complete the evaluation tasks.
On the BBH benchmark, Camel ran continuously for 266 hours and 49 minutes (approximately 11 days) without successfully completing the evaluation. Due to the excessive execution time and the absence of meaningful progress, the process was terminated manually. Similarly, for the GSM8K benchmark, Camel ran for approximately 73 hours and 23 minutes (around 3 days) but still failed to complete the evaluation tasks.
Our analysis indicates that uncontrolled context growth was the primary cause of these failures. Specifically, the framework attempted to build extremely large memory contexts and repeatedly shorten or reduce them to fit within the model’s context limit. This process caused frequent context overflow handling and message chunking. As a result, the system generated a large number of API calls and increased prompt-token usage while producing minimal or incomplete outputs. The evaluation process became computationally inefficient and was unable to reach task completion within a reasonable time frame.

As shown in Table \ref{tab:agentic_benchmark}, AutoGen achieved 90.94\% accuracy on the ARC benchmark, possibly because it consists of relatively simple tasks. However, on more complex benchmarks such as BBH and GSM8K, its accuracy fell to 24.08\% and 4.55\%, respectively, while its runtime remained significantly higher.
Our analysis indicates that AutoGen generated a large number of iterative agent interactions, resulting in repeated API calls for a single task. Complex benchmarks typically require multi-step reasoning, planning, and agent coordination, which increase prompt length and token consumption. As a result, the evaluation process produced a high volume of API requests, eventually exceeding the available API usage quota and causing the execution to terminate before completion. We observed a similar situation with the Mastra framework, which also generated a large number of API requests, and the execution failed due to exceeding the available API usage quota.

The Upsonic framework presented a different limitation related to computational cost. When running BBH and GSM8K on the same day, Upsonic consumed approximately 4,780,000 tokens within a single day, resulting in a cost of \$1,434.20. As a result of this unusually high cost, the evaluation process was stopped. During this period, the GSM8K benchmark was completed; however, BBH was not fully executed. 
The analysis indicates that the high computational cost is primarily due to very high prompt-token usage caused by repeated retries and uncontrolled context growth. The framework frequently caused extraction failures, which forced the system to repeatedly call the LLM until the output matched the expected format. In addition, the framework collects large contextual information through its memory and execution traces, where previous prompts, outputs, and task instructions are continuously added to the prompt context. As a result, the input context becomes significantly larger with each call. This behavior is reflected in the logs, which show extremely high input token counts per request, even though the final model responses remain relatively short. As a result, the combination of memory-based context growth, repeated extraction retries, and large prompt contexts considerably increased both token consumption and the overall API cost when running the Upsonic framework on the benchmark tasks.

\begin{tcolorbox}[width=\columnwidth, colback=gray!5, colframe=black!60, boxrule=0.5pt]
\textbf{Takeaway 1.}
Most frameworks showed mean accuracies between 74.57\% and 75.94\%, indicating a narrow performance plateau across architectures. Frameworks that fell outside this plateau underperformed, not because of weaker reasoning strategies but rather due to system-level orchestration issues such as poor memory management, unstable retry behavior, and uncontrolled context growth. This interpretation is further supported by the role-based multi-agent category, which contains both the best and some of the worst performers, with statistically significant within-category variance confirmed by ANOVA ($F(2,33)=16.61$, $p = 1.02 \times 10^{-5}$).
\end{tcolorbox}

\subsection{Performance Trade-Offs (RQ2)}
\label{Performance Trade-Offs (RQ2)}
In this section, we present the trade-off results of the agentic frameworks, which show the computational cost and execution time required to complete each task. The benchmarks contain a total of 16,495 tasks.
Across all experiments, the total number of input tokens consumed was 10.861 billion, while the total number of output tokens was 83,196. The large gap between input and output token consumption is due to the extensive input context and prompts required by the frameworks, whereas the outputs were usually very short, often limited to labels such as correct, incorrect, or failed. The overall experimental cost was \$3,154.30, with a total of 685,443 API requests. The total execution time was approximately 24 days (575 hours 37 minutes).

In the following sections, we present a detailed analysis of the trade-offs for each agentic framework. As shown in Table \ref{tab:framework_tradeoffs}, we report the total token consumption, the average number of tokens used per task across the three benchmarks, and the corresponding average cost in cents. This allows us to examine the computational cost associated with each framework. In addition, execution time is analyzed by reporting the total runtime and average time per task for each framework.

\begin{figure}
    \centering
    \includegraphics[width=1.0\linewidth]{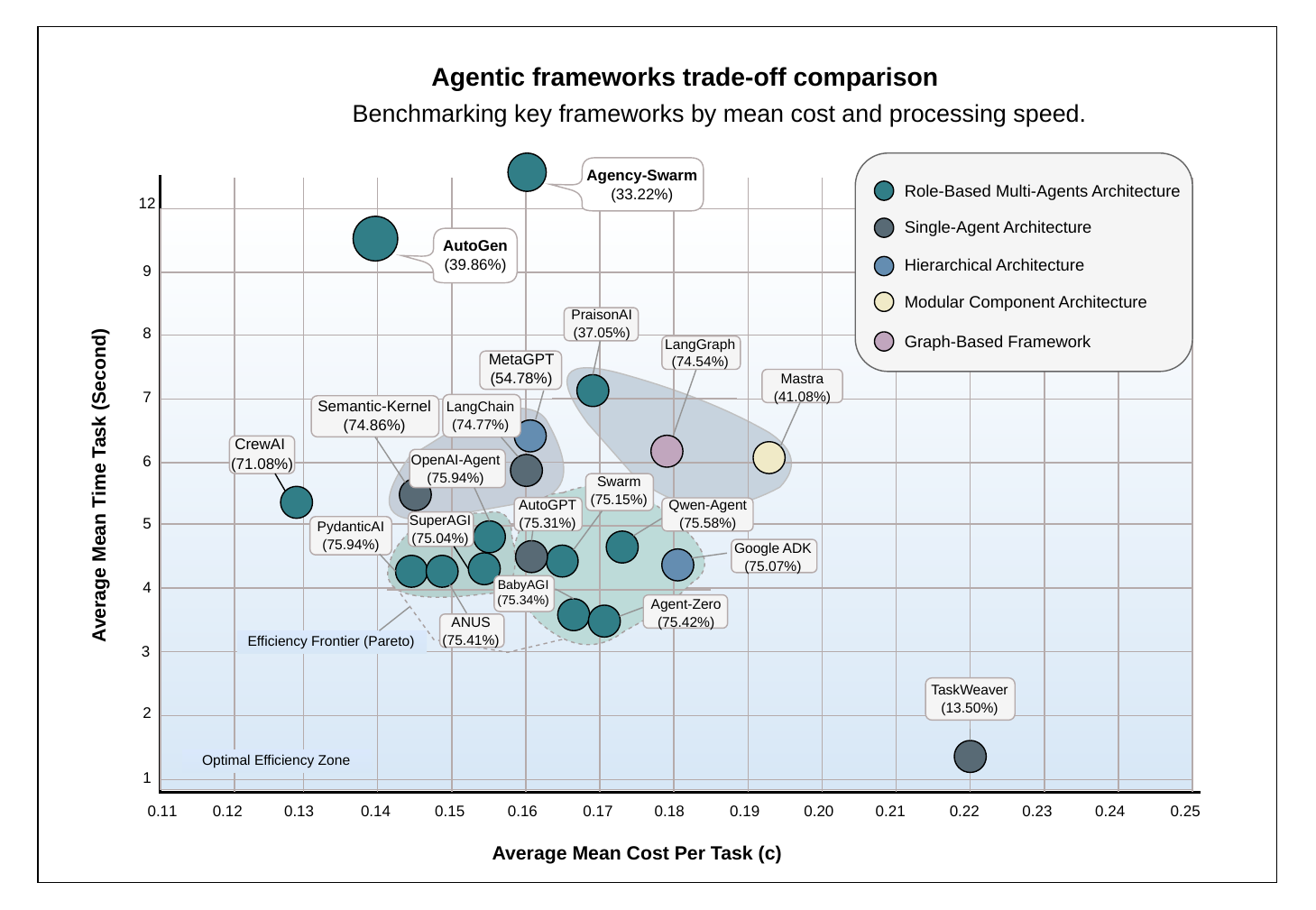}
    \caption{Performance–efficiency trade-off between average inference time and average token cost for the evaluated agentic frameworks.}
    \label{fig:tradsoff}
\end{figure}

Figure \ref{fig:tradsoff} presents a visual representation of the performance–efficiency trade-off for the selected agentic frameworks by plotting their average mean time per task (y-axis) against their average mean cost per task (x-axis) across the three benchmarks. We computed the mean response time per task for each framework across the three benchmarks, as well as the mean token consumption per task, which reflects the overall cost of execution. This analysis provides an overview of the time and cost efficiency of each framework when performing the benchmark tasks. In Figure \ref{fig:tradsoff}, different colors represent the categories of the agentic frameworks, while each data point corresponds to an individual framework. The figure also shows the mean accuracy of each framework, enabling a comprehensive comparison between reasoning performance, response time, and cost.

The results show that the average cost of most frameworks ranges between 0.14¢ and 0.18¢ per benchmark task, while the average response time typically varies between 4 and 6 seconds per task. As shown in Figure~\ref{fig:tradsoff}, four frameworks: Pydantic-AI, ANUS, SuperAGI, and OpenAI-Agent, fall within the efficient frontier zone, indicating highly efficient and cost-optimal performance. Their average completion time is around 4 seconds, with the average cost ranging from 0.14¢ to 0.15¢. Notably, all four belong to the role-based multi-agent category. Similarly, other frameworks from the same category, such as Swarm, Qwen-Agent, Agent-zero, and BabyAGI, also show efficient average completion times of between 3 and 4 seconds; however, their average cost is slightly higher, at around 0.17¢. CrewAI appears to be the most cost-efficient framework, with an average cost per task of 0.13¢, although its average completion time is slightly longer, at around 5 seconds. In contrast, frameworks such as AutoGen, Agency-Swarm, and PraisonAI exhibit slower performance, with average response times of 9 to 12 seconds per task, while their average cost consumption ranges between 0.14¢ and 0.17¢.

In addition, frameworks in the single-agent category vary in terms of both time and cost. For instance, AutoGPT, LangChain, and Semantic Kernel show an average cost per task of between 0.15¢ and 0.16¢, with average completion times ranging from 4 to 6 seconds. However, TaskWeaver demonstrates an unusual pattern: although its average completion time is lower than that of all other frameworks, at approximately 2 seconds, its average cost per task is comparatively higher, at 0.22¢. 

Frameworks in the hierarchical architecture category, such as Google ADK and MetaGPT, also show variation in time and cost. Google ADK demonstrates more time-efficient performance, with an average completion time of around 4 seconds, although its average cost is slightly higher at 0.18¢. In contrast, MetaGPT records an average completion time of around 6 seconds, while its average cost is lower at 0.16¢.

The modular architecture framework, Mastra, demonstrates a relatively low response time but results in the highest average cost per task (0.19¢) among the evaluated frameworks. Similarly, the graph-based framework, LangGraph, demonstrates an average response time of around 6 seconds per task, with an average cost of 0.18¢.

\textbf{Note}: 
During the experiments, several exceptional cases were observed. For example, the Camel framework required an unusually long time to complete tasks and often failed to finish execution. In contrast, Upsonic consumed a significantly larger number of tokens than the other frameworks. The AutoGen and Mastra frameworks resulted in repeated API calls for complex tasks, which significantly increased prompt length and token consumption. A detailed analysis of these four frameworks is provided in Section \ref{Observed Failures}, where we discuss the possible reasons for their failures based on an examination of their output datasets. 

\textbf{Limitation}:
One limitation of the dataset is that when an execution results in an error or failure, the number of tokens consumed is not recorded. Token information is only available for successful executions (both correct and incorrect), where the input and output token counts are reported. Therefore, in cases of failures or errors, we relied on the OpenAI usage dashboard to estimate the token consumption of the frameworks across the three benchmarks.

\begin{tcolorbox}[width=\columnwidth, colback=gray!5, colframe=black!60, boxrule=0.5pt]
\textbf{Takeaway 2.}
Most agentic frameworks maintain a suitable balance between reasoning performance, execution time, and cost, with average response times of 4–6 seconds per task and costs between 0.14¢ and 0.18¢ per task. However, higher computational cost and longer execution time do not necessarily result in better reasoning performance, highlighting the importance of efficient framework design and orchestration strategies.
\end{tcolorbox}

\clearpage
\newgeometry{left=1.8cm,right=1.8cm,top=1.8cm,bottom=1.8cm}
\begin{landscape}

\begin{table}[p]
\centering
\setlength{\tabcolsep}{6pt}
\renewcommand{\arraystretch}{1.9}

\resizebox{1.50\textheight}{!}{%
\begin{tabular}{|l|c|c|c|c|c|c|c|c|c|c|c|c|}

\hline
\textbf{Frameworks} & \multicolumn{4}{c|}{\textbf{BBH}} & \multicolumn{4}{c|}{\textbf{ARC}} & \multicolumn{4}{c|}{\textbf{GSM8K}} \\ \hline
 & \textbf{Avg-Time (s)} & \textbf{Total-Time (min)} & \textbf{Avg-Tok/Cost} & \textbf{Total-Tok/Cost}
 & \textbf{Avg-Time (s)} & \textbf{Total-Time (min)} & \textbf{Avg-Tok/Cost} & \textbf{Total-Tok/Cost}
 & \textbf{Avg-Time (s)} & \textbf{Total-Time (min)} & \textbf{Avg-Tok/Cost} & \textbf{Total-Tok/Cost} \\ \hline

\multicolumn{13}{|c|}{\textbf{Single-Agent Architecture}} \\ \hline
AutoGPT & 4.86s & 527 min & 1191/0.20¢ & 7760406/\$13.90 & 4.61s & 173 min & 696/0.12¢ & 1567793/\$2.86 & 3.32s & 413 min & 917/0.16¢ & 6850974/\$14.31 \\ \hline
TaskWeaver & 0.91s & 99 min & 1,362/0.23¢ & 8873320/\$15.84 & 2.05s & 77 min & 908/0.15¢ & 2045866/\$3.69 & 1.38s & 172 min & 2,252/0.39¢ & 16838592/\$29.84 \\ \hline
Semantic-kernel & 5.28s & 573 min & 1,176/0.20¢ & 7659727/\$13.72 & 5.86s & 220 min & 693/0.12¢ & 1560815/\$2.85 & 3.85s & 479 min & 908/0.159¢ & 6792003/\$12.25 \\ \hline
Langchain & 5.57s & 604 min & 916/0.16¢ & 7955670/\$14.24 & 6.08s & 228 min & 693/0.12¢ & 1560114/\$2.84 & 3.58s & 446 min & 916/0.16¢ & 6847193/\$12.35 \\ \hline

\multicolumn{13}{|c|}{\textbf{Role-Based Multi-Agent Architecture}} \\ \hline
BabyAGI & 4.39s & 476 min & 1,162/0.20¢ & 7572701/\$13.57 & 3.42s & 192 min & 1,041/0.18¢ & 2345503/\$4.22 & 3.28s & 409 min & 911/0.15¢ & 6813456/\$13.11 \\ \hline
Autogen & 9.31s & 1010 min & 974/0.17¢ & 6346009/\$11.42 & 10.34s & 388 min & 646/0.11¢ & 1454435/\$2.66 & 9.47s & 1179 min & 865/0.15¢ & 6461109/\$11.37 \\ \hline
Camel & \multicolumn{4}{c|}{Work stopped due to time limits} & 14.58s & 547 min & 781/0.13 & 1758031/\$3.26 & \multicolumn{4}{c|}{Work stopped due to time limits} \\ \hline
CrewAI & 6.01s & 652 min & 781/0.14¢ & 5085091/\$9.21 & 5.22s & 196 min & 627/0.10¢ & 1412455/\$2.59 & 6.53s & 813 min & 909/0.15¢ & 6793281/\$12.26 \\ \hline
SuperAGI & 5.06s & 549 min & 1,167/0.20¢ & 7604764/\$13.63 & 4.45s & 167 min & 693/0.12¢ & 1561377/\$2.85 & 3.55s & 442 min & 916/0.16¢ & 6845253/\$12.35 \\ \hline
Swarm & 4.39s & 476 min & 1,204/0.21¢ & 7840728/\$14.04 & 4.64s & 174 min & 697/0.12¢ & 1569081/\$2.86 & 3.67s & 457 min & 916/0.16¢ & 6847178/\$12.35 \\ \hline
Agency-swarm & 16.85s & 1829 min & 979/0.17¢ & 6378294/\$11.48 & 11.27s & 633 min & 901/0.15¢ & 2029749/\$3.67 & 8.53s & 1062 min & 924/0.16¢ & 6910939/\$12.46 \\ \hline
OpenAI-Agents-Python & 4.98s & 570 min & 1,156/0.20¢ & 7529003/\$13.49 & 4.91s & 184 min & 702/0.12¢ & 1582025/\$2.88 & 3.69s & 460 min & 921/0.16¢ & 6888398/\$12.42 \\ \hline
Agent-zero & 2.99s & 324 min & 1,016/0.17¢ & 6618517/\$11.90 & 3.42s & 192 min & 1,042/0.18¢ & 2346839/\$4.22 & 4.11s & 512 min & 914/0.16¢ & 6837492/\$12.33 \\ \hline
PraisonAI & 6.40s & 695 min & 932/0.16¢ & 6069847/\$10.94 & 10.29s & 386 min & 673/0.23¢ & 1515319/\$2.77 & 5.95s & 741 min & 808/0.14¢ & 6039478/\$10.94 \\ \hline
Qwen-Agent & 4.52s & 490 min & 1,165/0.20¢ & 7590411/\$13.60 & 4.82s & 181 min & 697/0.12¢ & 1570964/\$2.86 & 3.72s & 463 min & 920/0.16¢ & 6880026/\$12.41 \\ \hline
Pydantic-AI & 4.95s & 537 min & 1,141/0.19¢ & 7432928/\$13.32 & 5.68s & 213 min & 696/0.12¢ & 1568631/\$2.86 & 3.64s & 453 min & 914/0.16¢ & 6834467/\$12.33 \\ \hline
ANUS & 4.32s & 469 min & 907/0.15¢ & 7629524/\$13.67 & 4.59s & 172 min & 693/0.12¢ & 1561189/\$2.85 & 3.41s & 425 min & 907/0.15¢ & 6782353/\$12.24 \\ \hline

\multicolumn{13}{|c|}{\textbf{Hierarchical Architecture}} \\ \hline
MetaGPT & 7.85s & 852 min & 1273/0.22¢ & 8288503/\$14.82 & 4.42s & 166 min & 673/0.12¢ & 1488003/\$2.71 & 5.75s & 715 min & 873/0.15¢ & 6523929/\$12.64 \\ \hline
Google ADK & 4.77s & 518 min & 913/0.15¢ & 7588520/\$13.59 & 3.85s & 216 min & 1,047/0.18¢ & 2357361/\$4.24 & 3.39s & 422 min & 913/0.15¢ & 6829228/\$12.32 \\ \hline

\multicolumn{13}{|c|}{\textbf{Modular Component Architecture}} \\ \hline
Upsonic & \multicolumn{12}{c|}{Work stopped due to high cost} \\ \hline
Mastra & \multicolumn{4}{c|}{Failed} & 4.88s & 274 min & 1452/0.25¢ & 3270423/\$5.84 & 8.79s & 1094 min & 746/0.13¢ & 5577395/\$10.13 \\ \hline

\multicolumn{13}{|c|}{\textbf{Graph-Based Architecture}} \\ \hline
Langraph & 5.18s & 562 min & 921/0.16¢ & 7923272/\$14.18 & 4.91s & 184 min & 683/0.11¢ & 1538055/\$2.81 & 3.57s & 444 min & 921/0.16¢ & 6889649/\$12.42 \\ \hline
\end{tabular}
}
\caption{Trade-offs of agentic frameworks evaluated across the BBH, ARC, and GSM8K benchmarks, showing the average execution time per task, total execution time per benchmark, average cost per task, and total cost per benchmark for each framework.}
\label{tab:framework_tradeoffs}
\end{table}

\end{landscape}
\restoregeometry
\clearpage

\subsection{Multi-Benchmark Consistency (RQ3)}

Multi-benchmark consistency is examined in this section across all frameworks to evaluate how stable their performance remains in different reasoning tasks. 
First, we identify the absolute consistency in terms of benchmark performance, which measures the direct differences in accuracy across tasks. The results of this analysis are presented in Section~\ref{Absoulte Consistency}. We then examine relative (scale-adjusted) consistency, which accounts for differences in benchmark difficulty by normalizing performance variations. The findings of this analysis are discussed in Section~\ref{Relative (scale-adjusted) consistency}.

\subsubsection{Absolute Consistency}
\label{Absoulte Consistency}
As shown in Table \ref{tab:mad_total_deviation}, we used the mean absolute difference to measure the absolute consistency of each agentic framework across the three benchmarks (BBH, ARC, and GSM8K). Specifically, we computed the pairwise absolute differences between benchmark accuracies ($|BBH - ARC|$, $|BBH - GSM8K|$, and $|ARC - GSM8K|$) and combined them to obtain a total deviation score per framework. Lower deviation values indicate greater cross-benchmark stability, whereas higher values suggest performance variability across reasoning domains.

\begin{table*}[t]
\centering
\resizebox{\textwidth}{!}{%
\begin{tabular}{|l|c|c|c|c|}
\hline
\textbf{Framework} & \textbf{$|$BBH-ARC$|$} & \textbf{$|$BBH-GSM8K$|$} & \textbf{$|$ARC-GSM8K$|$} & \textbf{Total Deviation} \\ \hline
AutoGPT & 0.01 & 46.37 & 46.36 & 92.74 \\ \hline
TaskWeaver & 4.82 & 5.96 & 10.78 & 21.56 \\ \hline
Semantic-kernel & 1.78 & 44.67 & 46.45 & 92.90 \\ \hline
Langchain & 0.19 & 45.47 & 45.66 & 91.32 \\ \hline
BabyAGI & 2.01 & 45.29 & 47.30 & 94.60 \\ \hline
Autogen & 66.86 & 19.53 & 86.39 & 172.78 \\ \hline
CrewAI & 12.60 & 36.09 & 48.69 & 97.38 \\ \hline
SuperAGI & 0.38 & 45.60 & 45.98 & 91.96 \\ \hline
Swarm & 0.29 & 46.14 & 46.43 & 92.86 \\ \hline
Agency-swarm & 7.06 & 3.01 & 10.07 & 20.14 \\ \hline
OpenAI-Agents-Python & 1.39 & 46.75 & 48.14 & 96.28 \\ \hline
Agent-zero & 3.10 & 45.27 & 48.37 & 96.74 \\ \hline
PraisonAI & 66.19 & 15.64 & 81.83 & 163.66 \\ \hline
Qwen-Agent & 1.85 & 46.36 & 48.21 & 96.42 \\ \hline
Pydantic-AI & 1.96 & 46.27 & 48.23 & 96.46 \\ \hline
ANUS & 2.35 & 45.62 & 47.97 & 95.94 \\ \hline
MetaGPT & 18.85 & 72.75 & 91.60 & 183.20 \\ \hline
Google ADK & 0.34 & 45.92 & 45.58 & 91.84 \\ \hline
Langraph & 0.24 & 45.45 & 45.21 & 90.90 \\ \hline
\textbf{Mean Deviation} & 10.12 & 39.38 & 49.43 & -- \\ \hline
\end{tabular}%
}
\caption{Mean absolute differences between benchmark pairs and total deviation per framework. Lower total deviation indicates higher cross-benchmark stability.}
\label{tab:mad_total_deviation}
\end{table*}

The overall results highlight the fact that BBH and ARC show the strongest absolute consistency, with a mean deviation of 10.12, which is much lower than the deviations observed between BBH–GSM8K (39.38) and ARC–GSM8K (49.43). This finding indicates that agentic frameworks that perform well on BBH usually achieve similar performance on ARC. Since BBH includes logical, commonsense, hyperbaton, abstract, and general reasoning tasks, and ARC includes scientific reasoning tasks, this suggests that the frameworks generally perform better on these types of reasoning problems.
However, GSM8K shows larger deviations compared to both BBH and ARC, indicating that performance on mathematical word problems does not align as closely with performance on broader reasoning benchmarks. This pattern indicates that, while current agentic frameworks are reasonably effective in handling scientific, logical, commonsense, and abstract reasoning tasks, they struggle with mathematical reasoning tasks that require multi-step numerical computation and precise symbolic manipulation. 
This gap suggests that current frameworks lack mechanisms for reliable step-wise verification, numerical precision control, or specialized mathematical reasoning strategies. More details about the dataset can be found in our replication package \cite{rasheed_2026_19593242}.

\textbf{Note}: Camel, Upsonic, and Mastra were excluded from the absolute consistency analysis because valid results were not available for all three benchmarks. Since the calculation of mean deviation and total deviation requires complete benchmark results for each framework, these frameworks could not be included due to failure on one or two benchmarks.

\begin{table}[h]
\centering
\begin{tabular}{|c|l|c|c|c|}
\hline
\textbf{\# No} & \textbf{Framework} & \textbf{Mean Accuracy} & \textbf{Std Dev} & \textbf{CV} \\
\hline
1  & AutoGPT                & 75.31 & 26.77 & 0.355 \\ \hline
2  & TaskWeaver             & 13.50 & 5.40  & 0.400 \\ \hline
3  & Semantic-kernel        & 74.86 & 26.32 & 0.352 \\ \hline
4  & Langchain              & 74.77 & 26.31 & 0.352 \\ \hline

5  & BabyAGI                & 75.34 & 26.75 & 0.355 \\ \hline
6  & Autogen                & 39.86 & 45.30 & 1.137 \\ \hline
7  & CrewAI                 & 71.08 & 25.27 & 0.356 \\ \hline
8  & SuperAGI               & 75.04 & 26.44 & 0.352 \\ \hline
9 & Swarm                  & 75.15 & 26.72 & 0.356 \\ \hline
10 & Agency-swarm           & 33.22 & 5.17  & 0.156 \\ \hline
11 & OpenAI-Agents-Python   & 75.94 & 27.40 & 0.361 \\ \hline
12 & Agent-zero             & 75.42 & 27.08 & 0.359 \\ \hline
13 & PraisonAI              & 37.15 & 43.44 & 1.169 \\ \hline
14 & Qwen-Agent             & 75.85 & 27.32 & 0.360 \\ \hline
15 & Pydantic-AI            & 75.94 & 27.30 & 0.359 \\ \hline
16 & ANUS                   & 75.41 & 27.04 & 0.359 \\ \hline

17 & MetaGPT                & 54.78 & 48.37 & 0.883 \\ \hline
18 & Google ADK             & 75.07 & 26.41 & 0.352 \\ \hline


19 & Langraph               & 74.57 & 26.17 & 0.351 \\ \hline
\end{tabular}

\caption{CV across BBH, ARC, and GSM8K. Lower CV indicates higher proportional cross-benchmark stability.}
\label{tab:cv_consistency}
\end{table}

\subsubsection{Relative (Scale-Adjusted) Consistency}
\label{Relative (scale-adjusted) consistency}
To further quantify the proportional consistency of each framework across the benchmarks, we computed the CV, defined as $CV = \frac{\sigma}{\mu}$, where $\sigma$ represents the standard deviation of accuracies across BBH, ARC, and GSM8K, and $\mu$ denotes their mean. 

As shown in Table~\ref{tab:cv_consistency}, frameworks such as Agency-Swarm show the lowest CV (0.156), highlighting strong proportional stability across reasoning domains. In contrast, frameworks such as AutoGen (1.13), PraisonAI (1.16), and MetaGPT (0.88) exhibit substantially higher CV values, reflecting considerable performance variability across the benchmarks. Notably, the CV values for most frameworks fall approximately within the range of 0.35 to 0.36, suggesting that, although their absolute accuracies differ across the benchmarks, their proportional variability remains relatively similar. Overall, the CV analysis provides a scale-adjusted measure of cross-benchmark consistency, supporting the mean absolute deviation analysis by accounting for relative performance variation rather than absolute differences.

\textbf{Note}: Although CV provides a measure of relative variability, it should be interpreted cautiously in this study. Frameworks with low mean accuracy may show comparatively high CV values even when their absolute variation across benchmarks is modest. Likewise, a low CV does not necessarily indicate strong overall performance; for example, Agency-Swarm shows a relatively low CV but also low accuracy across the benchmarks. Therefore, CV should be interpreted in conjunction with mean accuracy and standard deviation. Additionally, Camel, Mastra, and Upsonic were excluded from the CV analysis because complete results were not available for all three benchmarks.

\begin{tcolorbox}[width=\columnwidth, colback=gray!5, colframe=black!60, boxrule=0.5pt]
\textbf{Takeaway 3.}
The results indicate that agentic frameworks show lower and less consistent performance on mathematical reasoning tasks than on other reasoning tasks. BBH and ARC remain relatively close, with a mean deviation of 10.12, whereas GSM8K shows much larger deviations of 39.38 from BBH and 49.43 from ARC. Furthermore, our analysis also highlights that MetaGPT, AutoGen, and PraisonAI have relatively high CV values, indicating greater proportional variation across benchmarks. For most frameworks, however, CV values fall between 0.35 and 0.36.
\end{tcolorbox}

\section{Discussion}
\label{Discussion}
This section discusses the key findings for each research question and outlines their implications for both academia and industrial practice. It highlights the performance and stability of agentic frameworks, the trade-offs between time efficiency and cost, and the challenges of achieving consistent accuracy across the evaluated benchmarks.

\subsection{Performance (RQ1)}
The most notable pattern in Table \ref{tab:agentic_benchmark} is not the identity of the top-ranked framework, but the limited differences in performance between most frameworks. 
Twelve frameworks, representing single-agent (AutoGPT, Semantic Kernel, LangChain), role-based multi-agent (BabyAGI, SuperAGI, Swarm, OpenAI Agents Python, Agent Zero, Qwen-Agent, PydanticAI, ANUS), hierarchical (Google ADK), and graph-based (LangGraph) categories, are separated by only 1.4 percentage points in mean accuracy (74.57–75.94\%). Their per-benchmark scores are highly similar: approximately 89–91\% on BBH, 44\% on GSM8K, and approximately 90–93\% on ARC. This pattern reflects not a meaningful ranking, but a performance plateau. The frameworks that fall outside the plateau do not fail because of weaker reasoning strategies. Instead, the common thread across every observed failure mode is memory and control-flow discipline. Uncontrolled context growth (Camel), unbounded retry on extraction failure (Upsonic), iterative agent chatter without termination criteria (AutoGen, Mastra)—these are engineering failures in how the frameworks manage state and bind their own behavior, not failures in how they reason. 


These findings are broadly consistent with prior work. As noted by Vaidhyanathan \textit{et al.} \cite{vaidhyanathan2025agentic}, agentic frameworks offer diverse approaches to solving complex tasks, but many remain at an early stage of maturity. This view is reinforced by the performance variation observed in our results, including the gap between high-performing specialized frameworks and lower-performing but widely adopted frameworks. In particular, limitations in memory mechanisms, error handling, security and privacy protection, and scalable deployment options may help explain why many frameworks still struggle to achieve reliable performance in real-world industrial settings \cite{garg2025designing, vaidhyanathan2025agentic}. To support the development of more mature frameworks for software engineering applications, future systems should incorporate stronger memory mechanisms, hybrid reasoning strategies, cost-effective designs, improved error recovery, and greater scalability. Addressing these limitations will be essential to making agentic AI frameworks more reliable and effective in real-world software engineering \cite{wei2026agentic}.

\textbf{Implications}: For researchers, these results point to several directions for improving the design and evaluation of agentic frameworks. (1) For the development of next-generation agentic frameworks, the observed convergence in performance should be interpreted as a challenge rather than rejection. These findings indicate that agent coordination strategies, role allocation mechanisms, and workflow orchestration techniques require further investigation to improve reasoning consistency and reliability. In addition, future frameworks may benefit from hybrid architectures that combine LLM-based reasoning with structured memory systems and symbolic or tool-based reasoning modules. Such developments could reduce computational overhead while enabling more reliable performance on complex real-world tasks. (2) The results also highlight the need for more comprehensive benchmarking methodologies and standardized evaluation frameworks for agentic systems. As LLM-based agent systems continue to evolve, large-scale empirical evaluations will remain essential for guiding framework design and improving reasoning reliability. (3) Finally, the public availability of the source code and replication package enables researchers to reproduce, validate, and extend this work. For example, future studies could explore the performance of agentic frameworks using open-source LLMs and compare their behavior with closed-source models. Future studies could also include newly developed frameworks and benchmarks to examine additional aspects such as privacy, security, robustness, and reliability. These factors are increasingly important for real-world deployment.

For practitioners, these results have several practical implications. (1) Industry is increasingly adopting agentic frameworks to develop AI-driven systems and autonomous workflows \cite{pati2025agentic}. In this context, this study provides practical guidance and evidence-based insights for practitioners in selecting suitable frameworks for real-world industrial deployment. (2) The proposed taxonomy enables practitioners to systematically identify frameworks that align with the specific requirements of their target application domain.

\subsection{Trade-off (RQ2)}
The results of this study highlight the trade-off between reasoning capability, computational cost, and execution time required by agentic frameworks to complete reasoning tasks. Our analysis indicates that the majority of frameworks are able to maintain reasonable execution efficiency while achieving acceptable reasoning performance. However, several notable exceptions highlight important inefficiencies: (1) Frameworks such as AutoGen, Agency Swarm, TaskWeaver, and Mastra exhibit higher execution time and cost compared to others. These frameworks also demonstrate lower reasoning accuracy, indicating that higher computational cost does not necessarily lead to better performance. (2) The Camel framework showed high runtime due to uncontrolled context growth, where the framework attempted to construct extremely large memory contexts that repeatedly exceeded the model’s context limit. This led to frequent context overflow handling and message chunking, which increased the number of API calls and token usage while producing minimal outputs. (3) Upsonic consumed significantly higher costs due to repeated LLM calls when outputs failed to match the required extraction format, and accumulated memory and execution traces expanded the prompt context. These observations indicate that framework design, orchestration strategies, and agent coordination mechanisms play a critical role in determining the efficiency and effectiveness of agentic systems.

\textbf{Implications}: For researchers, these results point to several important avenues of research: (1) Although many frameworks demonstrate promising reasoning capabilities, the observed trade-offs between performance, cost, and execution time suggest that current agentic architectures are still far from optimal in terms of computational efficiency. Future research should therefore focus on developing more resource-efficient agent orchestration mechanisms, improved task decomposition strategies, and optimized reasoning pipelines that reduce computational overhead while maintaining strong reasoning performance. (2) It is also important to incorporate cost and efficiency metrics into the evaluation of agentic frameworks, as reasoning accuracy alone may not provide a complete understanding of framework performance in real-world scenarios. Overall, achieving a balance between reasoning capability, computational efficiency, and execution time remains a key challenge in designing agentic frameworks.

For practitioners, the findings provide several important implications: (1) These findings offer empirical guidance for selecting frameworks that ensure reliable reasoning performance while minimizing computational cost, which is crucial for large-scale deployments where API usage, token consumption, and inference time directly affect operational expenses and system scalability. (2) The results indicate that higher computational cost and longer execution time do not necessarily guarantee better reasoning performance. Therefore, practitioners should carefully evaluate agentic frameworks before large-scale deployment to ensure efficient and reliable system performance. In real-world deployments, organizations must balance reasoning performance with computational efficiency and operational cost, particularly when agentic systems operate at scale. (3) Some frameworks, such as Agent-zero and ANUS, demonstrate better cost efficiency and lower execution time while maintaining competitive reasoning performance, whereas several widely used frameworks require higher execution time and computational resources. This implies that practitioners should carefully consider framework efficiency in addition to popularity when selecting agentic frameworks for real-world deployment.
In summary, selecting frameworks that balance accuracy, execution speed, and cost efficiency is critical.

\subsection{Consistency (RQ3)}

The results of this study highlight the consistency of agentic frameworks across different reasoning benchmarks. Specifically, the BBH and ARC benchmarks demonstrate relatively consistent performance, with a mean deviation of 10.12, indicating that frameworks tend to perform similarly across these two reasoning tasks. This suggests that many agentic frameworks are capable of maintaining stable reasoning performance when evaluated on benchmarks that focus on logical, commonsense, and abstract reasoning tasks. However, when comparing the outcomes of BBH and ARC with the GSM8K benchmark, a significantly higher deviation was observed. The mean deviation between BBH and GSM8K was 39.38, while the deviation between ARC and GSM8K was 49.43. These larger deviations indicate that the reasoning performance of agentic frameworks becomes considerably less consistent when evaluated on mathematical reasoning tasks. 

This finding further reinforces the observation that current agentic frameworks are generally more reliable in knowledge-based and logical reasoning tasks, but struggle to maintain consistent performance in numerical and multi-step mathematical reasoning problems. Similar observations have been reported in previous studies, for instance \cite{mirzadeh2024gsm, zhang2025comprehension, hosseini2024not, yuan2023scaling, tan2025scaling}. These studies suggest that language models tend to perform more consistently on logical and commonsense reasoning tasks, while showing greater variability and lower reliability on mathematical reasoning benchmarks because of limitations in multi-step numerical reasoning and arithmetic generalization. These studies further highlight that LLMs struggle with mathematical reasoning due to their dependence on pattern recognition rather than symbolic reasoning. As a result, they often fail to generalize to unseen arithmetic tasks and frequently make errors in multi-step numerical calculations. Collectively, these findings  suggest that current LLMs lack the robust algorithmic reasoning capabilities required for reliable mathematical problem solving \cite{zhang2024careful}.

\textbf{Implications}: For researchers, the significant deviation observed in mathematical reasoning benchmarks suggests the need for new agent architectures and reasoning mechanisms that better support structured and multi-step numerical problem solving.

For practitioners, these findings suggest that agentic frameworks show stable and predictable performance in applications that primarily involve logical, commonsense, and knowledge-based reasoning tasks, making them suitable for domains such as information analysis, research assistance, and decision support systems. However, the findings also suggest that organizations, particularly in domains that require precise numerical reasoning, financial analysis, or complex quantitative decision-making, should apply additional safeguards such as verification mechanisms or hybrid architectures, due to the current limitations of agentic frameworks in mathematical reasoning tasks. Additionally, the findings indicate that organizations should consider task–domain alignment when selecting agentic frameworks, ensuring that the chosen framework demonstrates consistent performance on benchmarks that closely resemble the intended real-world application scenarios.

\section{Threats to Validity}

\label{Threats to Validity}
We follow the guidelines proposed by Wohlin \textit{et al}.~\cite{wohlin2012experimentation} to identify and discuss potential threats to validity. This section deals with four categories: internal, external, construct, and conclusion validity.

\textbf{Internal validity} refers to how well the experiments were conducted and the extent to which the resulting data can be trusted.
First, a potential threat to internal validity arises from the hardware configuration used to conduct the experiments. All evaluations were executed on a fixed workstation with a specific CPU, memory capacity, storage configuration, and overall compute resources. These hardware characteristics may affect performance-related measurements such as execution time, latency, throughput, and the occurrence of timeouts or execution instability during agent workflows. As a result, observed differences in efficiency- or cost-related metrics may be partially influenced by the execution environment rather than by the design of the agentic frameworks alone. To reduce this threat, all frameworks were executed within the same controlled hardware and software environment, ensuring consistent compute resources, operating system configuration, and runtime dependencies across experiments. To ensure methodological consistency and reduce other influencing factors, we applied identical framework configurations and decoding parameters across all evaluated agentic frameworks. While this controlled setup supports fair comparison, it may limit the exploration of how different frameworks behave under alternative decoding strategies. Investigating the stability of agentic framework performance under different decoding configurations will therefore be addressed in future work.

Another potential threat to internal validity arises from the dependence of agentic frameworks on LLMs, which are known to exhibit non-deterministic behavior and output variability for the same input. Such inconsistency may influence reasoning outcomes and lead to changes in performance across repeated runs. To reduce this threat, each reasoning task was executed multiple times, and the same input was evaluated three times under identical settings. The final results were derived by aggregating the outcomes across these runs (e.g., averaging accuracy scores), reducing random variation. 
Finally, the last threat to internal validity arises from the use of static benchmarks in this study. Although existing reasoning benchmarks provide controlled and reproducible evaluation environments, they are primarily based on static datasets. Such benchmarks may not fully capture the emerging capabilities of agentic systems, such as adaptive reasoning, continuous learning, and long-horizon task execution. This limitation is also reflected in the choice of benchmark versions. For example, we used the BBH benchmark, whereas a more advanced version, BBEH, was introduced subsequently with more challenging reasoning tasks. However, at the time our experiments were conducted, BBEH was not yet available, and therefore BBH was the most suitable benchmark we could use. As a result, benchmark-based evaluation may only partially represent the behavior of agentic systems in dynamic real-world environments. This limitation is further reinforced by the current lack of standardized benchmarks specifically designed to evaluate agentic systems in complex real-world settings involving dynamic context, long task horizons, and iterative decision-making processes. Future work should therefore incorporate newer and more challenging benchmarks to provide a more comprehensive assessment of agentic frameworks.

\textbf{External validity} concerns the extent to which the findings of this study can be generalized beyond the evaluated settings. A potential threat to external validity arises from the evolving nature of agentic frameworks. The selection of frameworks in this study was based on a systematic search conducted on GitHub starting in July 2025. As a result, some agentic frameworks that were released or gained prominence after this period were not included in the evaluation. Therefore, the set of analyzed frameworks may not fully represent the most recent developments in the rapidly evolving agentic framework landscape.
In addition, we employed a few-shot prompting approach combined with CoT to evaluate the reasoning capabilities of the agentic frameworks. However, this approach may not fully capture the external validity of real-world software engineering scenarios. In practice, reasoning is often an iterative and interactive process where developers provide incremental context, respond to model hallucinations, and offer corrective feedback—elements that are not fully represented by the static examples used in a few-shot setting.
Therefore, although benchmark-based experiments enable controlled and reproducible comparisons, they may only approximate the interactive reasoning processes that occur in real-world development environments.

\textbf{Construct validity} addresses whether the evaluation accurately measures the intended outcomes of the benchmark-based comparison of agentic frameworks. In this study, a potential threat to construct validity arises from the use of a single LLM (GPT-5.2) to evaluate the agentic frameworks. Agentic frameworks may indicate different behaviors, interaction patterns, or performance characteristics when paired with different LLMs. Although evaluating multiple models could provide a broader understanding of framework behavior, this was not feasible in the current study due to cost constraints. Specifically, the API usage cost for GPT-5.2 in this project exceeded \$3154.30. In future work, we plan to extend the evaluation to five additional models, including open-source alternatives, in order to assess whether the observed findings remain consistent across different underlying LLMs.
Another potential threat to construct validity arises from the training data limitations of the selected LLM used by agentic frameworks. These models are trained on fixed and potentially outdated datasets, which may limit their ability to reason about recent or evolving information. To mitigate this threat, we enabled external environment support, such as \textit{web search} tools, allowing agents to access up-to-date information from the internet during task execution. This setup helps reduce the impact of static training data and better reflects realistic agent usage scenarios; however, the extent to which external tools influence reasoning behavior may vary across frameworks.

\textbf{Conclusion validity} refers to the extent to which the obtained results support the conclusions derived from the study. In this study, agentic framework executions may be affected by runtime issues such as timeouts, resource exhaustion, or partial failures, especially when complex reasoning tasks are executed or when frameworks involve multiple agent interactions. Such issues may lead to incomplete results for certain frameworks on specific benchmarks, which may introduce bias into the comparative analysis.
To mitigate this risk, we applied consistent execution constraints across all frameworks, including limits on token generation and execution time where necessary. All failed or incomplete runs were systematically documented and excluded uniformly across frameworks to avoid skewing the comparisons.

Another limitation relates to abnormal execution behavior observed in a small number of frameworks. During the experiments, some frameworks generated excessive context growth, repeated agent interactions, or repeated API retries, which resulted in unusually long runtimes or extremely high token consumption. For example, one framework consumed a very large number of tokens within a short execution period, resulting in substantial API cost before the experiment was manually terminated to prevent uncontrolled resource usage. Moreover, when a framework execution failed or terminated unexpectedly, detailed token usage information was not always recorded by the logging pipeline. In such cases, token usage was estimated using the OpenAI usage dashboard rather than the experiment logs. As a result, the reported token statistics may slightly underestimate the true computational cost for frameworks that experienced execution failures or incomplete runs. These limitations should therefore be considered when interpreting the comparative efficiency and cost analysis.

\section{Conclusion}
\label{conclusion}
In this study, we systematically selected and evaluated 22 agentic frameworks (see Table~\ref{tab:agentic_taxonomy}) to examine their capabilities in reasoning tasks. The study was structured around the three research questions defined in Section~\ref{Research Questions}. To address these questions, we conducted experiments on three widely used benchmarks: BBH, ARC, and GSM8K. Our evaluation focused on three key aspects. First, we compared the overall performance of the selected frameworks by analyzing their mean accuracy across the benchmarks. Second, we investigated the trade-offs associated with each framework by measuring inference time and computational cost in order to understand their resource requirements. Third, we analyzed the consistency of framework performance across different benchmark datasets to assess their stability in diverse reasoning scenarios. Based on the experimental results, the main findings of this study are summarized as follows:

\begin{itemize}
    \item The findings highlight that most of the evaluated frameworks achieved relatively similar mean accuracy across the selected benchmarks. However, a smaller group of frameworks performed worse, primarily because of orchestration-related issues, such as uncontrolled context growth, retry loops, and inefficient agent interactions, rather than limitations in reasoning itself.

    \item The results indicate a clear trade-off between performance, response time, and operational cost. While many frameworks converged within a narrow accuracy range, they still varied in latency and API cost, indicating that framework selection should not be based on accuracy alone but also on efficiency and cost considerations.

    \item The findings further show that mathematical reasoning remains a major limitation across all evaluated frameworks. In particular, performance on GSM8K was lower than on BBH and ARC, suggesting that current agentic frameworks largely inherit the underlying model’s weaknesses in multi-step numerical reasoning rather than overcoming them.

    \item Overall, the study indicates that the practical value of agentic frameworks for reasoning-intensive tasks depends more on orchestration robustness, including memory management, failure handling, and cost control, than on the architectural category alone.
\end{itemize}

The findings of this study provide useful insights for both researchers and practitioners. For researchers, the results highlight several directions for future investigation into the design and evaluation of agentic frameworks. In addition, the source code and replication data are publicly available, enabling other researchers to reproduce our experiments and extend this work by evaluating agentic frameworks with multiple LLMs. Future studies could also incorporate newly developed agentic frameworks and recently proposed benchmarks to examine additional aspects such as privacy, security, robustness, and reliability, which are becoming increasingly important for real-world deployment of agentic systems. For practitioners, this study offers empirical guidance for selecting appropriate frameworks for industrial applications by considering their reasoning performance, computational trade-offs, and overall consistency across different tasks.

\section*{Acknowledgment}

This project was co-funded by the BF/MAISA/SW kust project (2024–2026) and the ANSE-BF research project (2025–2027), both funded by Business Finland. These projects represent a collaboration between academia and eight leading Finnish companies. They managed all the API costs and other expenses required for the completion of this project. 

\section*{Data Availability}

To support the replication and validation of this study, we have publicly released the source code on GitHub \cite{agenticframeworkcomparison}. In addition, the complete benchmark results for all evaluated frameworks, including benchmark data files, execution logs, and the Excel file used for data analysis, are provided in Zenodo \cite{rasheed_2026_19593242}. The released materials are publicly available to facilitate further research and future extensions of this work.

\section*{Declaration of AI Assistance}

During the preparation of this manuscript, the authors used language models to assist with grammar refinement, sentence restructuring, and formatting improvements. Following the use of these tools, the authors carefully reviewed and revised the content and assume full responsibility for the final version of the publication.


\bibliography{sn-bibliography}

\end{document}